\documentclass[letterpaper]{article} 
\usepackage{aaai2026}  
\usepackage{times}  
\usepackage{helvet}  
\usepackage{courier}  
\usepackage[hyphens]{url}  
\usepackage{graphicx} 
\usepackage{natbib}  
\usepackage{caption} 
\usepackage{algorithm}
\usepackage{algorithmic}
\usepackage{subfigure}
\usepackage{booktabs}
\usepackage{multirow}
\usepackage{amsthm}
\newtheorem{proposition}{Proposition}

\newtheorem{corollary}{Corollary}

\newtheorem{remark}{Remark}

\usepackage{subcaption}
\usepackage{array}
\usepackage{makecell}
\usepackage{enumitem}
\usepackage{newfloat}
\usepackage{listings}
\DeclareCaptionStyle{ruled}{labelfont=normalfont,labelsep=colon,strut=off} 
\floatstyle{ruled}
\newfloat{listing}{tb}{lst}{}
\floatname{listing}{Listing}
\usepackage{amsmath}  
\usepackage{amssymb}  

\title{FedTopo: Topology-Informed Representation Alignment in \\ Federated Learning under Non-I.I.D. Conditions}
\author{
    Ke Hu\textsuperscript{\rm 1},
    Liyao Xiang\textsuperscript{\rm 2},
    Peng Tang\textsuperscript{\rm 1,3}\thanks{Corresponding author.},
    Weidong Qiu\textsuperscript{\rm 1,3}
}
\affiliations{
    \textsuperscript{\rm 1}School of Computer Science, Shanghai Jiao Tong University\\
    \textsuperscript{\rm 2}John Hopcroft Center for Computer Science, Shanghai Jiao Tong University\\
    \textsuperscript{\rm 3}Shanghai Key Laboratory of Integrated Administration Technologies for Information Security\\

    crestiny@sjtu.edu.cn,
    xiangliyao08@sjtu.edu.cn,
    tangpeng@sjtu.edu.cn,
    qiuwd@sjtu.edu.cn
}

\usepackage{bibentry}

\begin{document}

\maketitle

\begin{abstract}
Current federated-learning models deteriorate under heterogeneous (non-I.I.D.) client data, as their feature representations diverge and pixel- or patch-level objectives fail to capture the global topology which is essential for high-dimensional visual tasks. We propose \textbf{FedTopo}, a framework that integrates \textbf{Topological-Guided Block Screening (TGBS)} and \textbf{Topological Embedding (TE)} to leverage topological information, yielding coherently aligned cross-client representations by \textbf{Topological Alignment Loss (TAL)}. 

First, Topology-Guided Block Screening (TGBS) automatically selects the most topology-informative block, i.e., the one with maximal topological separability, whose persistence-based signatures best distinguish within- versus between-class pairs, ensuring that subsequent analysis focuses on topology-rich features. Next, this block yields a compact Topological Embedding, which quantifies the topological information for each client. Finally, a Topological Alignment Loss (TAL) guides clients to maintain topological consistency with the global model during optimization, reducing representation drift across rounds. 

Experiments on Fashion-MNIST, CIFAR-10, and CIFAR-100 under four non-I.I.D. partitions show that FedTopo accelerates convergence and improves accuracy over strong baselines. Code is available in Supplementary Materials.

\end{abstract}

\section{Introduction}

Federated Learning (FL) has emerged as a powerful paradigm for decentralized training across multiple clients. However, one of its central challenges lies in the presence of data heterogeneity---commonly referred to as the \emph{non-I.I.D.} problem. When local datasets are skewed, each client’s gradient points toward a different optimum, and performance of the aggregated global model decreases. From a representational perspective, skewed local data acts analogously to having different teachers emphasize distinct aspects of a concept. Each client learns a different interpretation of the data, leading to discrepancies in the learned feature representations. Recent studies~\shortcite{ye2023fedfm} have shown that this divergence is a key manifestation of non-I.I.D. data.

\begin{figure*}[htbp]
  \centering
  \begin{minipage}[b]{0.28\textwidth}
    \centering
    \includegraphics[width=\linewidth]{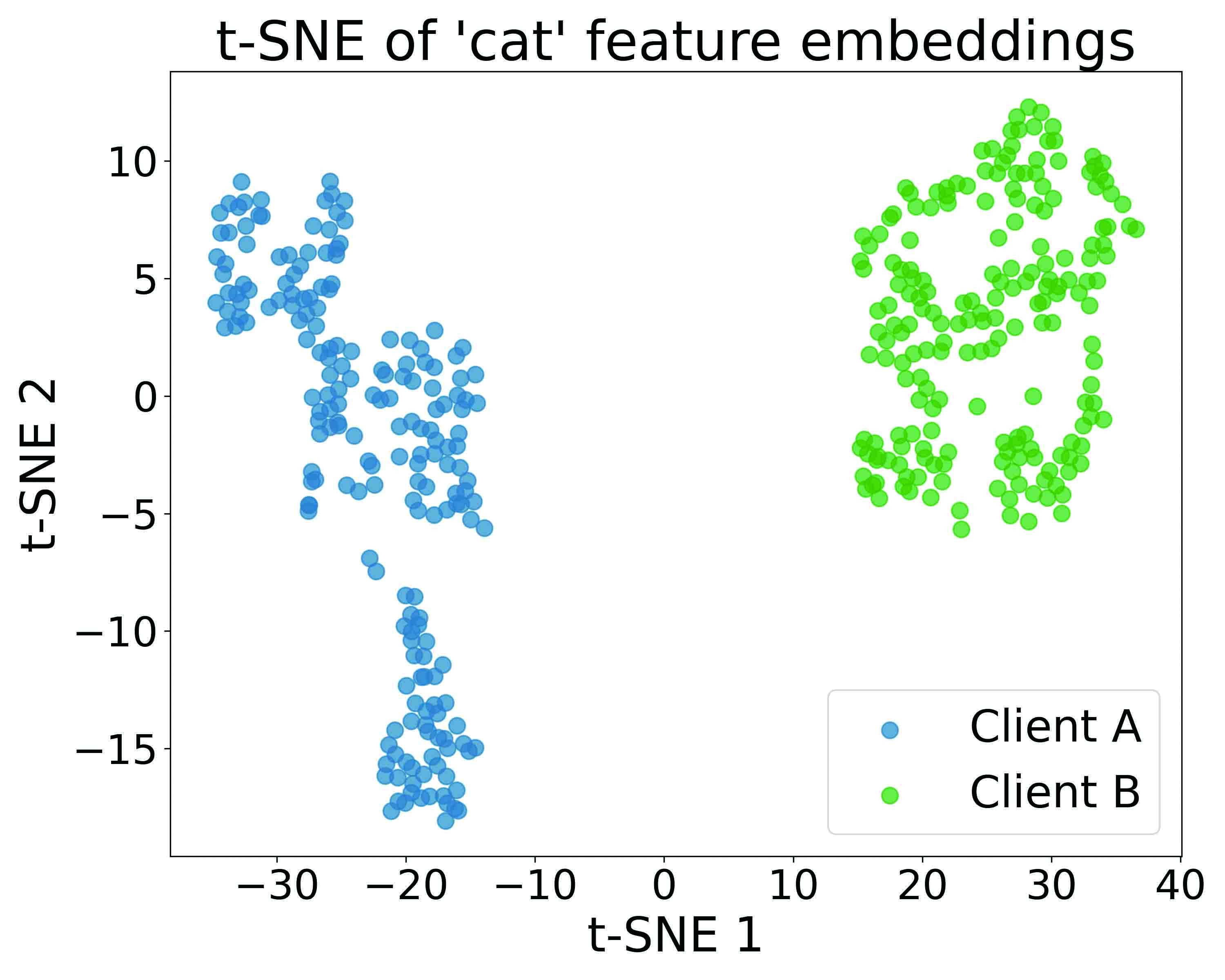}
    \par\vspace{0.5ex}
    \centerline{(a)}
  \end{minipage}%
  \hfill
  \begin{minipage}[b]{0.44\textwidth}
    \centering
    \includegraphics[width=\linewidth]{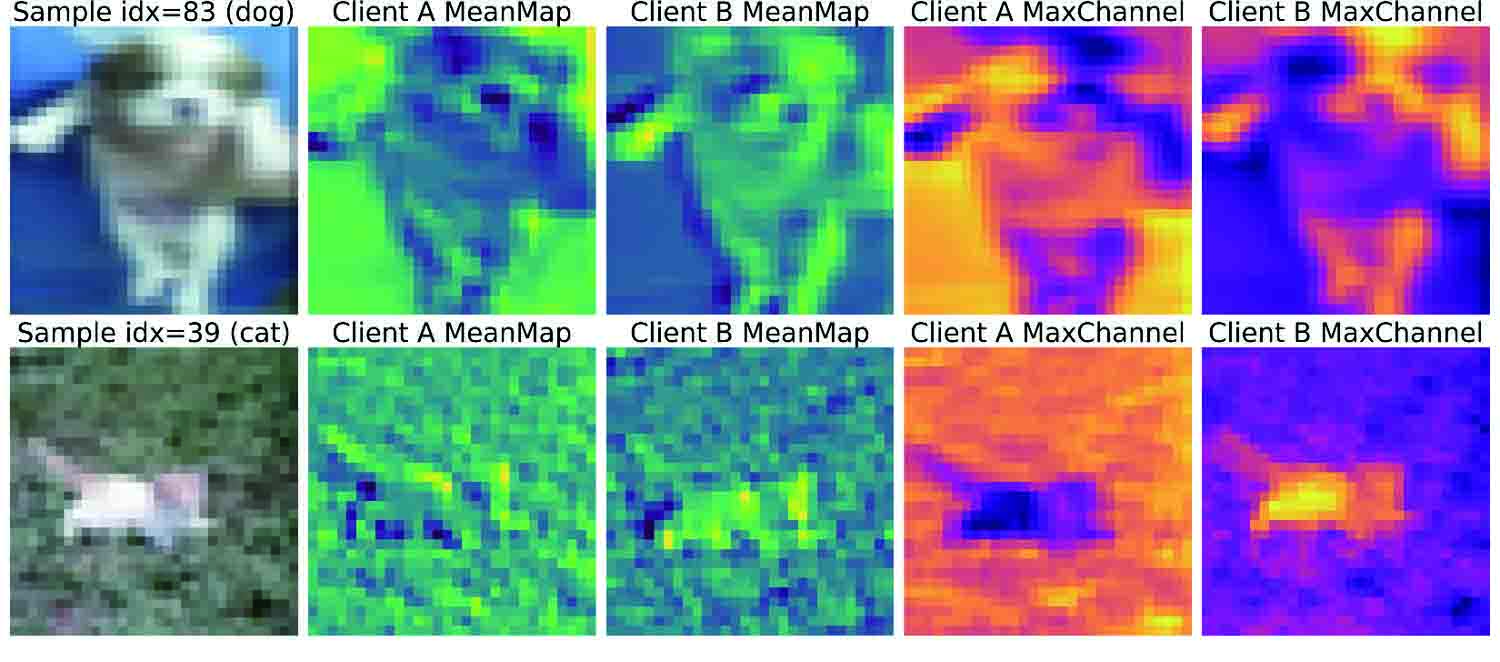}
    \par\vspace{2.0ex}
    \centerline{(b)}
  \end{minipage}%
  \hfill
  \begin{minipage}[b]{0.28\textwidth}
    \centering
    \includegraphics[width=\linewidth]{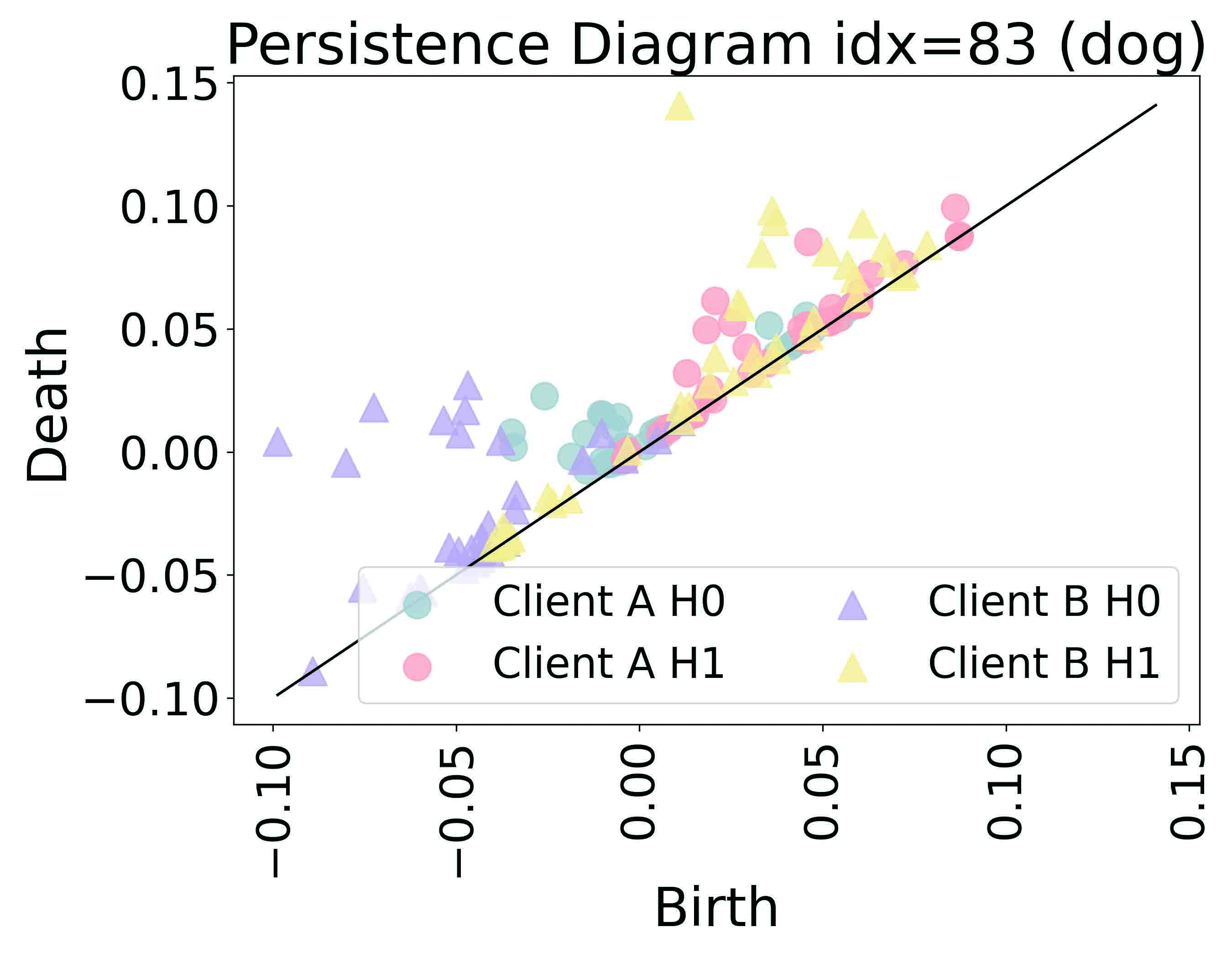}
    \par\vspace{0.5ex}
    \centerline{(c)}
  \end{minipage}
  \caption{
    Comparison of feature distributions, activation maps, and topological structure.
    (a) t-SNE of 'cat' feature embeddings.~~
    (b) Feature/activation map visualizations for two samples.~~
    (c) Persistence diagrams for the selected dog sample.
  }
  \label{fig:intro}
\end{figure*}

To intuitively illustrate this issue, we conduct a toy experiment using the CIFAR-10 dataset. We select two clients: Client~A holds samples of cats and dogs, while Client~B has disjoint cat samples and additional ship samples. Figure~\ref{fig:intro} illustrates feature inconsistency across clients from three perspectives: (1) \textbf{Feature distribution} (Figure~\ref{fig:intro}(a)): We extract intermediate representations of ``cat'' samples and project them using t-SNE. The resulting clusters are client-specific with minimal overlap, indicating inconsistent representations even for the same class. (2) \textbf{Activation maps} (Figure~\ref{fig:intro}(b)): For two selected inputs (``dog'' and ``cat''), we visualize the mean and max-channel activations. The maps reveal noticeable differences in spatial focus and intensity across clients. (3) \textbf{Topological structure} (Figure~\ref{fig:intro}(c)): For a selected dog image (index 83), we compute persistence barcodes and diagrams. Discrepancies in H$_0$/H$_1$ topological features suggest mismatched topological summaries of the learned features.
This example demonstrates that non-I.I.D. client datasets induce misalignment not only in parameter space but also in the deeper representational geometry and topology of feature representations~\shortcite{ye2023fedfm}. However, current methods generally ignore the latter.

\subsection{Contributions}

We propose \textbf{FedTopo}, a topology-aware federated learning framework that enforces cross-client topological consistency to mitigate representation drift in non-I.I.D. scenarios. Our key contributions are:

\begin{itemize}
    \item \textbf{Topology-Guided Block Screening (TGBS)}: A layer selection mechanism that identifies the block with maximal topological class separability, shown to correspond to the highest mutual information with class labels.

    \item \textbf{Topological Embedding (TE)}: A compact, persistence-based representation derived from the selected block, which is provably Lipschitz-stable under input and model perturbations.

    \item \textbf{Topological Alignment Loss (TAL)}: A regularizer that aligns each client's TE with that of the global model. We provide convergence analysis and introduce an adaptive weighting scheme based on a proposed non-I.I.D. metric and the temporal dynamics of topological loss.

\end{itemize}

FedTopo achieves both theoretical soundness and empirical effectiveness, consistently outperforming strong baselines on FMNIST, CIFAR-10, and CIFAR-100 under diverse non-I.I.D. conditions.

\section{Related Work}

\subsection{Non-I.I.D. Federated Learning}

Federated Learning (FL) enables collaborative training across decentralized clients while preserving data privacy~\shortcite{iqua2025federated}. However, performance degrades significantly under non-I.I.D.\ data due to representational drift and unstable convergence~\shortcite{mdpi2025federated}. To address this, recent methods focus on improving the alignment between local and global models~\shortcite{iqua2025federated}. For instance, FedCCFA~\shortcite{neurips2024fedccfa} clusters client classifiers and aligns intermediate features to mitigate concept drift. FedSA~\shortcite{arxiv2025fedsa} introduces semantic anchors to regularize representation learning and alleviate feedback loops caused by statistical and model heterogeneity. A detailed summary of recent non-I.I.D.\ FL methods is provided in Appendix~\ref{appendix:related-work}.

\subsection{Topological Data Analysis (TDA)}

TDA provides global, multi-scale descriptors that are robust to noise, deformation, and transformation---properties well suited to the challenges of non-I.I.D.\ FL. Unlike pixel-level metrics~\shortcite{zia2024topological}, topological invariants such as connected components and loops remain stable under continuous perturbations~\shortcite{adams2015persistence}. Prior work incorporates persistent homology into segmentation~\shortcite{clough2019explicit, clough2022topological}, reconstruction, and classification tasks, yielding improved robustness and generalization~\shortcite{hu2021topoloss, ma2024beyond, barannikov2021representation}. These results suggest that topological structure can provide complementary supervision in heterogeneous learning environments.

\subsection{Topological Alignment in Deep Learning}

Topological alignment has been used to promote consistent representations across distributions and modalities. In graph learning, aligning higher-order structures---such as edge orbits---enhances embedding robustness~\shortcite{sun2022htc, chen2023cgcl}. Other work shows that neural networks can implicitly preserve topological invariants such as connectivity and loops~\shortcite{halverson2025topinv}. In domain adaptation and change detection, explicitly aligning topological structure improves geometric consistency and classification boundaries~\shortcite{czolbe2021topodiff}. These findings motivate our approach of aligning topological descriptors across clients to reduce representation drift in federated learning.

\section{Preliminaries}

FedTopo uses topological descriptors to summarize the topological information in feature maps across heterogeneous clients. This section introduces the topological tools for converting feature maps into compact topological descriptors, including cubical filtrations, persistence diagrams (PDs), and persistence images (PIs).

\subsection{Feature Maps to Topological Descriptors}

Figure~\ref{fig:te_pipeline} illustrates the full computation pipeline using a CIFAR-10 example.

\subsubsection{Cubical Filtrations and Persistence Diagrams}

Given a local sample $x \in \mathcal{D}_i$, let $\phi_\ell(x; \mathbf{w}) \in \mathbb{R}^{C \times H \times W}$ denote the activation tensor at a selected layer $\ell$ of a local model $f(\cdot; \mathbf{w})$. We extract $n_C$ channels, each yielding a 2D activation map $A \in \mathbb{R}^{H \times W}$ (Figure~\ref{fig:te_pipeline}(a)).

To analyze the topology of $A$, we apply \textbf{cubical persistent homology} using a \textbf{sublevel set filtration}:
\begin{equation}
A^\lambda = \left\{ (i,j) \in [H] \times [W] \;\middle|\; A_{i,j} \leq \lambda \right\}.
\label{eq:sublevel-filtration}
\end{equation}
As $\lambda$ increases, more pixels are included, and topological features—such as components (H$_0$) and holes (H$_1$)—emerge and vanish. This produces a \textbf{persistence diagram} (Figure~\ref{fig:te_pipeline}(b)) that records their birth and death thresholds $(b_i, d_i)$.

\subsubsection{From Persistence Diagrams to Persistence Images}\label{sec:pi}

Persistence diagrams are expressive but difficult to use directly due to their non-Euclidean nature and variable size. To enable compatibility with learning systems, we adopt the \textbf{persistence image (PI)}~\shortcite{adams2017persistence}, a fixed-size, differentiable representation.

\begin{figure*}[htbp]
\centering
\includegraphics[width=1.0\textwidth]{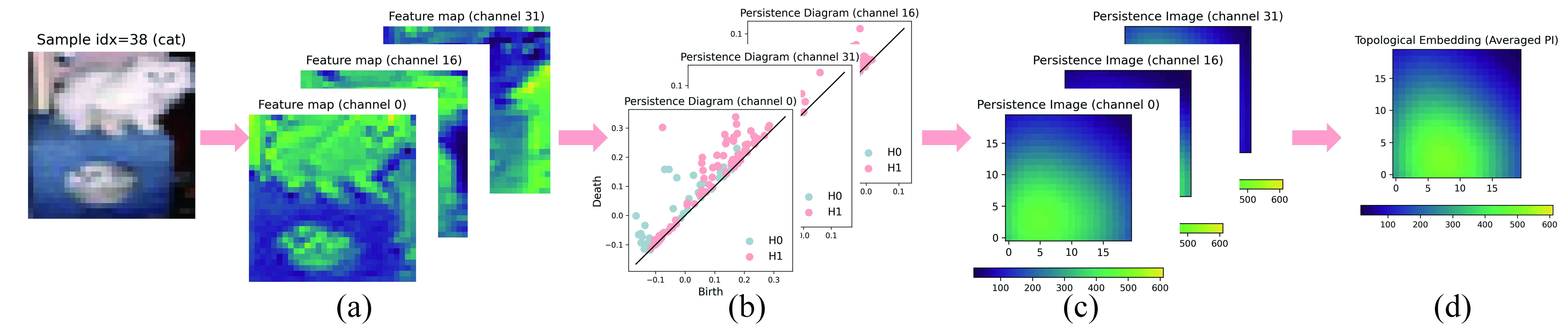}
\caption{Topological embedding pipeline illustrated using a CIFAR-10 \texttt{cat} image. We show the input image, its intermediate feature map (a), the persistence diagrams from selected channels (b), the corresponding persistence images (c), and the final averaged topological embedding (d).}
\label{fig:te_pipeline}
\end{figure*}

Each diagram is mapped from birth–death $(b_i, d_i)$ to birth–persistence  $(b_i, p_i = d_i - b_i)$ coordinates. A Gaussian kernel is then applied to each point to create a smooth density:
\begin{equation}
\rho(x, y) = \sum_i \exp\left( -\frac{(x - b_i)^2 + (y - p_i)^2}{2\sigma^2} \right),
\label{eq:persistence-image-density}
\end{equation}
which is rasterized onto a fixed grid and flattened into a vector $\psi_c(x; \mathbf{w}) \in \mathbb{R}^M$, the PI for channel $c$ (Figure~\ref{fig:te_pipeline}(c)). In Section~\ref{sec:te}, we aggregate these across channels to obtain the final \emph{Topological Embedding}.

\section{Proposed Method}

We organize this section around the three components of FedTopo. First, we identify the most topology-informative block via Topology-Guided Block Screening (TGBS). Second, we extract a compact Topological Embedding (TE) from that block as a topological descriptor to represent each client's feature topology. Third, we introduce the Topological Alignment Loss (TAL) with an adaptive $\alpha$ scheduler to align local and global topologies and mitigate representation drift.

\subsection{Topology-Guided Block Screening (TGBS)} \label{sec:tgbsmethod}

\subsubsection{Motivation}

We hypothesize that a block's topological signature, if it consistently distinguishes within-class and between-class samples, carries strong semantic signals. To target alignment where it matters most, FedTopo applies TGBS to identify the \emph{most topology-informative block}—i.e., the one with highest topological class separability.

\subsubsection{TGBS Design}
TGBS identifies the most topologically discriminative block by measuring its topological class separability (see Algorithm~\ref{alg:TGBS}). For each candidate block, we extract activations, apply PCA for channel reduction, compute cubical persistence diagrams (H$_0$ and H$_1$), and convert them into fixed-size Persistence Images (PIs) for comparison (see Section~\ref{sec:pi} for details). We then calculate Euclidean distances~\shortcite{adams2017persistence} between PIs and evaluate their ability to distinguish same-class vs.\ different-class sample pairs using ROC-AUC. 

The block with the highest AUC is selected as topology-informative. In ResNet-18, TGBS selects \texttt{layer2} (AUC 0.814) as the most separable layer. Selections for other architectures are listed in Appendix~\ref{appendix:method-details}.

\begin{algorithm}[tb]
\caption{TGBS: Automatic Selection of the Topology-Informed Block}
\label{alg:TGBS}
\textbf{Input}: Backbone $f$, candidate blocks $\mathcal{B}$, validation set $\mathcal{D}$, set of distance metrics $\mathcal{M}$, number of pairs $N_{\text{pairs}}$\\
\textbf{Output}: Topology-informative block $b^\star$ (and AUC score for each $b$)
\begin{algorithmic}[1]
\FOR{each block $b \in \mathcal{B}$}
    \STATE Extract features $\{\mathbf{z}_i\} \gets f_b(\mathcal{D})$
    \STATE Obtain embedding $\{\tilde{\mathbf{z}}_i\} \gets \textsc{DimReduce}(\{\mathbf{z}_i\})$
    \STATE Compute persistence diagrams $\{\mathrm{PD}_i\} \gets \textsc{PersistentHomology}(\{\tilde{\mathbf{z}}_i\})$
    \FOR{each metric $m \in \mathcal{M}$}
        \STATE Randomly sample $N_{\text{pairs}}$ within-class pairs $\mathcal{P}_{\mathrm{w}}$ and $N_{\text{pairs}}$ between-class pairs $\mathcal{P}_{\mathrm{b}}$
        \FOR{each $(j,k) \in \mathcal{P}_{\mathrm{w}} \cup \mathcal{P}_{\mathrm{b}}$}
            \STATE Compute distance $d_{jk} \gets \textsc{Dist}(\mathrm{PD}_j, \mathrm{PD}_k; m)$
            \STATE Set similarity $s_{jk} \gets -d_{jk}$
        \ENDFOR
        \STATE Calculate $\mathrm{AUC}_m(b) \gets \textsc{ROC\_AUC}(\{s_{jk}\})$
    \ENDFOR
    \STATE Compute average $\mathrm{AUC}(b) \gets \frac{1}{|\mathcal{M}|} \sum_{m} \mathrm{AUC}_m(b)$
\ENDFOR
\STATE $b^\star \gets \arg\max_{b \in \mathcal{B}} \mathrm{AUC}(b)$
\STATE \textbf{return} $b^\star$ and $\{\mathrm{AUC}(b)\}_{b \in \mathcal{B}}$
\end{algorithmic}
\end{algorithm}

\subsection{Topological Embedding (TE)} \label{sec:te}

Given the per-channel persistence images $\psi_c$ obtained in Preliminaries~\S\ref{sec:pi}, we build a single descriptor for each sample by averaging over channels:
\begin{equation}
\mathbf{t}_i(x; \mathbf{w}) = \mathcal{T}_\ell(x; \mathbf{w}) = \frac{1}{K} \sum_{c=1}^{K} \psi_c(x; \mathbf{w})
 \in \mathbb{R}^M.
\label{eq:topo-embedding}
\end{equation}
Here, $K$ is the number of channels, and the resulting vector $\mathcal{T}_\ell$ is a compact topological descriptor of the activation at layer $\ell$. Channel averaging removes permutation noise and reduces the dimension from $C \times H \times W$ to $M$, yielding a communication-efficient representation that underpins our Topological Alignment Loss (Section~\ref{sec:tal}).

\paragraph{Computation cost.}
FedTopo introduces lightweight computation overhead. The most expensive step—computing topological embeddings via cubical persistent homology—has a worst-case time complexity of $O(C(HW)^{\omega})$ per layer, with $C$ channels, spatial size $H \times W$, and $\omega \le 3$ (e.g., $\omega \approx 2.37$ for modern algorithms). In practice, efficient 2D cubical implementations with union-find and Morse heuristics yield near-linear runtime $\tilde O(CHW)$. For example, \texttt{layer2} in ResNet-18 ($C=128$, $H=W=8$) involves $\sim 2.9 \times 10^4$ cells per block, incurring only a minor cost per batch as FedTopo requires just one additional forward of the selected block.

\paragraph{Communication cost.}
FedTopo does not add communication overhead during training, as all topological computations are local. If a single topological embedding of dimension $M$ is uploaded per round (e.g., for monitoring), the overhead is
\begin{equation}
\Delta_{\text{FedTopo}} = \frac{M}{|\mathbf{w}|}.
\end{equation}
With $M = 64$, this corresponds to $5.8 \times 10^{-6}$ for ResNet-18 ($|\mathbf{w}| \approx 11$M) and $6.4 \times 10^{-4}$ for a small CNN ($|\mathbf{w}| \approx 0.1$M), both well below $0.1\%$ of model size. Thus, FedTopo preserves the communication efficiency of FedAvg.

\subsection{Topological Alignment Loss (TAL) with Adaptive Weighting} \label{sec:tal}

The local objective for client $i$ combines \textbf{cross-entropy} and \textbf{TAL}, which compares the client’s descriptor to the global reference:

\begin{align}
\mathcal{L}_i(\mathbf{w}_i; \bar{\mathbf{w}}) &=
\underbrace{\mathbb{E}_{(x, y) \sim \mathcal{D}_i} \left[ \mathrm{CE}(f(x; \mathbf{w}_i), y) \right]}_{\text{Classification term}} \notag \\
&\quad + \alpha \, \underbrace{\mathbb{E}_{x \sim \mathcal{D}_i} \left[ \left|\mathbf{t}_i(x; \mathbf{w}_i) - \mathbf{t}_i(x; \bar{\mathbf{w}})\right|^2 \right]}_{\text{TAL}},
\label{eq:tal-loss}
\end{align}

where $\bar{\mathbf{w}}$ denotes the previous global model, and $\alpha > 0$ controls the regularization strength. The objective in Eq.~\eqref{eq:tal-loss} aligns local and global topological embeddings, reducing representational drift across clients.

\paragraph{Adaptive Scheduling of $\alpha$.}
To modulate regularization dynamically, we set $\alpha_i = \lambda^{(r,e)} \cdot \alpha_i^{\text{base}}$, where $\lambda^{(r,e)}$ depends on global round and local epoch, and $\alpha_i^{\text{base}}$ reflects the current topology loss. We consider four scheduling strategies: (1) Warm-up, a static baseline without feedback; (2) Linear-Topo, which linearly scales $\alpha_i$ with the smoothed TAL loss; (3) Piecewise, which adjusts $\alpha_i$ using a piecewise-linear response to the loss; and (4) Smooth-Topo, which applies EWMA smoothing and a softened transformation for improved stability. See Appendix~\ref{app:alpha-scheduling} for details.

\subsection{System Overview of FedTopo}

FedTopo integrates three components into a unified topology-aware federated learning framework: (1) Topology-Guided Block Screening (TGBS) to identify the most topology-informative block, (2) construction of a compact Topological Embedding (TE) to describe the topology of feature maps, and (3) Topological Alignment Loss (TAL) with adaptive $\alpha$ scheduling to align local and global topological structures. Together, these components reduce representational drift and improve generalization under non-I.I.D.\ conditions.

\begin{figure*}[t]
\centering
\includegraphics[width=0.95\textwidth]{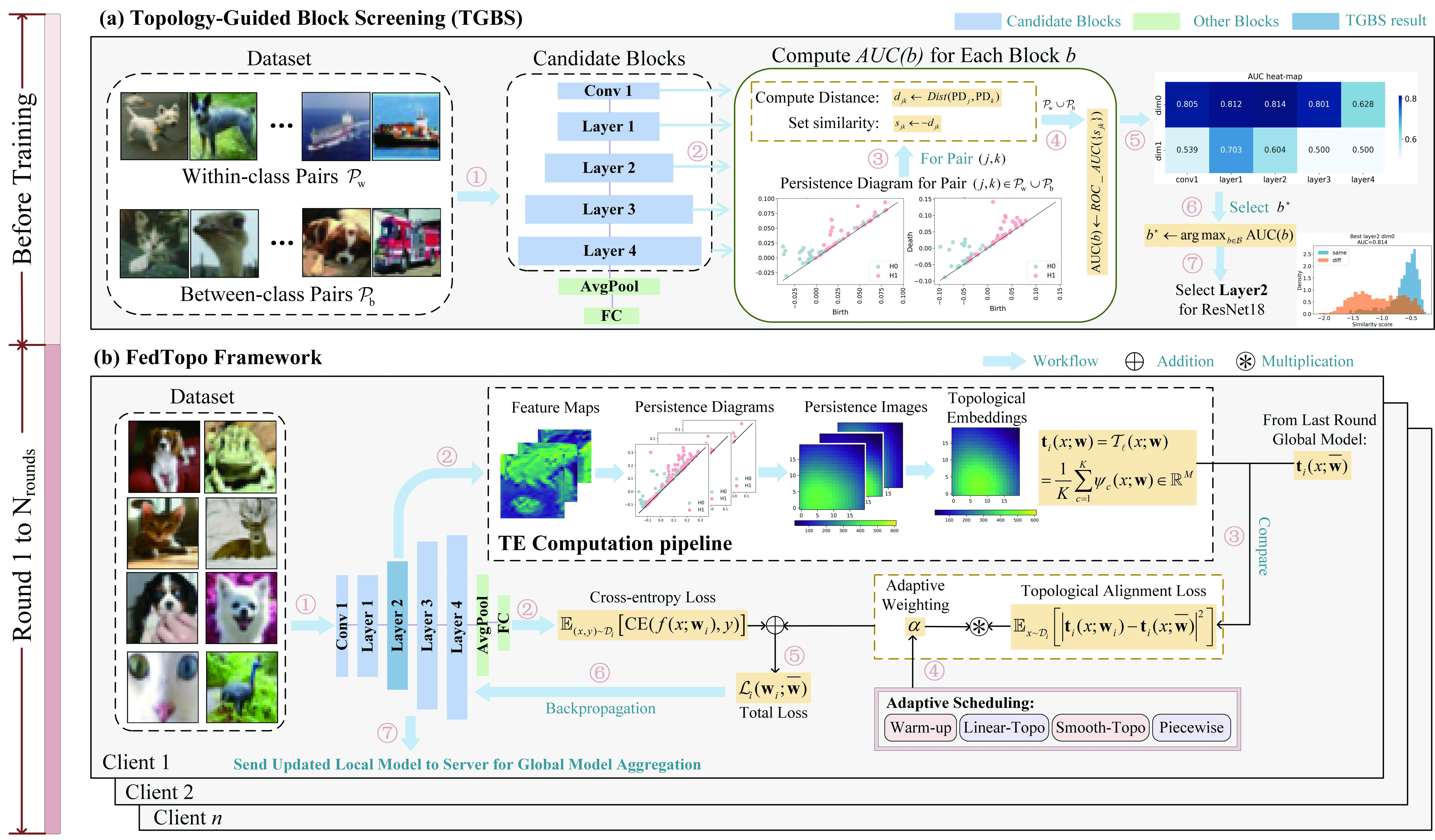}
\caption{System overview of \textbf{FedTopo}, illustrated using ResNet-18 and CIFAR-10. (a) Before training, TGBS selects the most topology-informative block based on class separability in topological space. (b) During training, each client computes topological embeddings from intermediate features and applies Topological Alignment Loss (TAL) with adaptive scheduling.}
\label{fig:framework}
\end{figure*}

Figure~\ref{fig:framework} illustrates the overall pipeline (Algorithm~\ref{alg:fedtopo}). In the pre-training stage (Figure~\ref{fig:framework}a), TGBS selects the block with the highest class separability in topological space. During each communication round (Figure~\ref{fig:framework}b), clients extract intermediate features from the selected block and compute the corresponding TE following the procedure described in Preliminaries. TAL is then applied alongside standard cross-entropy loss to update the local model. The adaptive $\alpha$ scheduler adjusts the strength of topological regularization across rounds to balance task performance and topological consistency.

\begin{algorithm}[tb]
\caption{FedTopo: Topology-Aware Federated Learning}
\label{alg:fedtopo}
\textbf{Input}: Initial global model $\bar{\mathbf{w}}^{(0)}$, client data $\{\mathcal{D}_i\}$, number of rounds $R$, TGBS-selected block $\ell^\star$ \\
\textbf{Output}: Final global model $\bar{\mathbf{w}}^{(R)}$
\begin{algorithmic}[1]
    \STATE \textbf{Preprocessing (server-side):}
    \STATE \quad Run TGBS on validation data to select $\ell^\star$ (Algorithm~\ref{alg:TGBS})
    \FOR{round $r = 1$ to $R$}
        \STATE Server broadcasts $\bar{\mathbf{w}}^{(r-1)}$ to all clients
        \FOR{each client $i$ in parallel}
            \STATE Initialize local model $\mathbf{w}_i^{(r)} \leftarrow \bar{\mathbf{w}}^{(r-1)}$
            \FOR{local epoch $e = 1$ to $E$}
                \STATE Sample minibatch $(x, y) \sim \mathcal{D}_i$
                \STATE Compute cross-entropy loss $\mathcal{L}_{\text{CE}}$
                \STATE Compute $\mathbf{t}_i(x; \mathbf{w}_i)$ and $\mathbf{t}_i(x; \bar{\mathbf{w}}^{(r-1)}) (Eq.~\eqref{eq:topo-embedding})$
                \STATE Compute TAL $\mathcal{L}_{\text{TAL}}$ using Eq.~\eqref{eq:tal-loss}
                \STATE Update $\alpha_i$ using adaptive scheduler
                \STATE Update $\mathbf{w}_i$ via gradient step on total loss $\mathcal{L}_{\text{CE}} + \alpha_i \mathcal{L}_{\text{TAL}}$ using Eq.~\eqref{eq:tal-loss}
            \ENDFOR
            \STATE Client sends updated $\mathbf{w}_i^{(r)}$ to server
        \ENDFOR
        \STATE Server aggregates client models to update $\bar{\mathbf{w}}^{(r)}$
    \ENDFOR
    \STATE \textbf{return} final model $\bar{\mathbf{w}}^{(R)}$
\end{algorithmic}
\end{algorithm}

\subsection{Theoretical Guarantees}

We present key theoretical justifications for the design of FedTopo. Formal proofs of all propositions are provided in Appendix~\ref{appendix:theory}.

\subsubsection{AUC as a Proxy for Mutual Information} \label{sec:auc-mi}

TGBS selects the block with the highest between-class separability, measured by AUC. We formally justify this by showing that AUC is monotonic with mutual information:

\begin{proposition}[AUC Implies Mutual Information]
Let $T_b = T_b(X)$ be the topological embedding at block $b$, and define the distance distributions:
\begin{align}
d_b^+ &\sim \left\{ \|T_b(x) - T_b(x')\| \;\big|\; Y(x) = Y(x') \right\}, \notag\\
d_b^- &\sim \left\{ \|T_b(x) - T_b(x')\| \;\big|\; Y(x) \neq Y(x') \right\}.
\label{eq:auc-dist}
\end{align}

Then AUC is defined as:
\begin{equation}
\mathrm{AUC}_b = \Pr_{u \sim d_b^+,\, v \sim d_b^-} \left[ u < v \right].
\label{eq:auc-def}
\end{equation}

If $\mathrm{AUC}_{b_1} > \mathrm{AUC}_{b_2}$, then $I(T_{b_1}; Y) > I(T_{b_2}; Y)$, implying

\begin{equation}
\arg\max_b \mathrm{AUC}_b = \arg\max_b I(T_b; Y).
\label{eq:auc-mi}
\end{equation}
\label{prop:auc-mi}
\end{proposition}

\subsubsection{Lipschitz Stability of Topological Embedding}

TE is computed via channel-wise Persistence Images (PIs), which are known to be Lipschitz continuous~\shortcite{adams2017persistence, carriere2017sliced}. This ensures the TE inherits input stability:

\begin{proposition}[Stability of Topological Embedding]
Let $\mathbf{t}_i(x; \mathbf{w})$ be the TE from $\phi_\ell(x; \mathbf{w})$. Then,
\begin{equation}
\left\| \mathbf{t}_i(x; \mathbf{w}) - \mathbf{t}_i(x'; \mathbf{w}) \right\|_2 \leq L \cdot \|x - x'\|_2.
\label{eq:lipschitz-te}
\end{equation}
\end{proposition}

This guarantees robustness of TAL to small input perturbations and supports stable convergence of the TAL loss.

\subsubsection{Convergence Guarantee via Proximal Regularization}

We further show that TAL acts as a proximal regularizer, satisfying FedProx-like convergence properties~\shortcite{li2020fedprox, yuan2022fedprox}:

\begin{proposition}[FedProx-style Convergence under TAL]
If $F_i(w)$ is $L$-smooth and the TAL term is $\mu$-strongly convex, then the global objective converges to an $\varepsilon$-stationary point in $O\left( \frac{L}{\mu} \log\frac{1}{\varepsilon} \right)$ gradient steps.
\end{proposition}

\section{Experiments}

We empirically evaluate FedTopo from multiple perspectives: overall performance across diverse non-I.I.D.\ settings, visualization of representation alignment, and ablation studies assessing the contributions of key components including TGBS, TAL, and adaptive $\alpha$ scheduling.

\subsubsection{Data-partition schemes}

We consider four non-I.I.D.\ partition types: quantity skew, label skew, noise-based skew, and fixed-$k$ label assignment.

\paragraph{Amplitude noise skew (N-skew).}
Each client shares the same inputs $\{x_i\}$ but adds Gaussian noise $n_{i,j} \sim \mathcal{N}(0, \sigma_j^2 I)$. Here, $\sigma_j = \frac{j-1}{n-1} \bar{\sigma}$ and $\sigma_n = 0$, so the client's data becomes $\mathcal{D}_j = \{ (x_i + n_{i,j}, y_i) \}$.

\paragraph{Label skew (L-skew).}
For each class $c$, we normalize the distribution $p_{c,j}^{(\ell)}$ to obtain $\hat{p}_{c,j} = \frac{p_{c,j}^{(\ell)}}{\sum_{m=1}^n p_{c,m}^{(\ell)}}$. The sample indices for client $j$ are defined as $I_{c,j} = \left[ \sum_{m<j} \hat{p}_{c,m} N_c, \sum_{m\le j} \hat{p}_{c,m} N_c \right)$, and the dataset is $\mathcal{D}_j = \bigcup_c \{ (x_i, y_i) \mid i \in I_{c,j} \}$.

\paragraph{Fixed-$k$ label skew.}
Each client is assigned $k$ labels from $[K]$ so that all labels are covered. For class $c$, the indices $\mathcal{I}_c$ are split into shards $\{\mathcal{I}_c^{(s)}\}$, and each client $j$ is assigned data from classes in $\mathcal{C}_j$, yielding $\mathcal{D}_j = \bigcup_{c \in \mathcal{C}_j} \mathcal{I}_c^{(j, c)}$.

\paragraph{Quantity skew (Q-skew).}
For each client $C_j$, we sample proportions $\mathbf{p}^{(q)} \sim \operatorname{Dir}_n(\alpha_q)$ and normalize them as $\hat{p}_j = p_j^{(q)} / \sum_{m=1}^n p_m^{(q)}$. The sample indices for client $j$ are then assigned by $I_j = \left[ \sum_{m<j} \hat{p}_m N, \sum_{m\le j} \hat{p}_m N \right)$, resulting in $\mathcal{D}_j = \{ (x_i, y_i) \mid i \in I_j \}$.

Full definitions are provided in Appendix~\ref{appendix:experimental-details}.

\subsubsection{Baselines}

We compare FedTopo with three families of federated learning methods designed for non-I.I.D.\ data:
(1) \textbf{Aggregation regularization and momentum-based methods:} FedAvg~\shortcite{mcmahan2017communication}, FedAvgM~\shortcite{li2020federated}, FedNova~\shortcite{wang2020tackling}, FedProx~\shortcite{li2020federated}, SCAFFOLD~\shortcite{karimireddy2020scaffold}, and FedDyn~\shortcite{jin2023feddyn}. These methods modify the aggregation process using regularization, correction terms, or momentum to mitigate client drift and accelerate convergence.
(2) \textbf{Representation alignment and personalization methods:} MOON~\shortcite{li2021model}, FedFTG~\shortcite{zhang2022fine}, FedPAC~\shortcite{xu2023personalized}, FedFM~\shortcite{ye2023fedfm}, FedCCFA~\shortcite{neurips2024fedccfa}, and FedSA~\shortcite{arxiv2025fedsa}. These approaches promote consistency in feature representations or personalize the model to accommodate client-specific distributions.
(3) \textbf{Server-side knowledge transfer methods:} FedDF~\shortcite{lin2020ensemble}, FedDC~\shortcite{gao2022feddc}, and FedGen~\shortcite{zhu2021data}. These techniques rely on distillation or synthetic data generation at the server to reduce cross-client distribution shift.

\subsubsection{Implementation Details}

We use mini-batch SGD~\shortcite{robbins1951stochastic} for local optimization with $E = 5$ local epochs. All methods use a batch size of 32 and learning rate of 0.01. We train with $n = 5$ clients and run 10 communication rounds. For feature extraction, \textbf{FedTopo} uses \texttt{conv1} for the CNN model on MNIST and FMNIST, and \texttt{layer2} for ResNet-18 on CIFAR-10. Additional hyperparameters are listed in Appendix~\ref{appendix:experimental-details}.

\subsection{Performance Evaluation}

\subsubsection{Accuracy and Convergence}

We run each method five times and report the mean global test accuracy in Table~\ref{tab:main_accuracy}. FedTopo outperforms all baseline methods across all datasets and non-I.I.D.\ partitioning schemes. On FMNIST, CIFAR-10, and CIFAR-100, FedTopo achieves the best accuracy, improving over the second-best method by up to 8.69\%, 11.05\%, 8.44\% respectively.
These results demonstrate that FedTopo effectively addresses the challenges of non-I.I.D.\ data, reducing representation drift and improving feature alignment across clients.

\begin{table*}[htbp]
  \centering
  \begin{tabular}{c|cccc|cccc|cccc}
    \toprule
    \textbf{Dataset} & \multicolumn{4}{c|}{FMNIST} & \multicolumn{4}{c|}{CIFAR-10} & \multicolumn{4}{c}{CIFAR-100} \\
    \midrule
    \textbf{Non-I.I.D.} & N & L & K & Q & N & L & K & Q & N & L & K & Q \\
    \midrule
    FedAvg & 60.01 & 83.83 & 75.97 & 91.07 & 65.10 & 86.47 & 66.69 & 90.62 & 46.87 & 60.53 & 49.35 & 63.43 \\
    FedAvg-M & 65.43 & 87.70 & 80.21 & 92.86 & 67.87 & 89.76 & 66.85 & 93.39 & 48.87 & 62.83 & 49.47 & 65.37 \\
    FedNova & 63.19 & 85.28 & 73.85 & 90.11 & 64.90 & 86.07 & 66.80 & 87.73 & 46.73 & 60.25 & 49.43 & 61.41 \\
    FedProx & 66.68 & 87.49 & 75.91 & 90.57 & 63.47 & 85.85 & 66.99 & 89.35 & 45.70 & 60.10 & 49.57 & 62.55 \\
    SCAFFOLD & 68.78 & 87.43 & 80.35 & 90.22 & 63.33 & 86.88 & 69.91 & 86.03 & 45.60 & 60.82 & 51.73 & 60.22 \\
    FedDyn & 63.47 & 89.68 & 83.95 & 90.33 & 65.10 & 86.47 & 68.32 & 90.62 & 46.87 & 60.53 & 50.56 & 63.43 \\
    \midrule
    MOON & 63.93 & 88.95 & 74.76 & 90.19 & 54.45 & 76.30 & 67.74 & 81.73 & 38.23 & 53.41 & 50.13 & 57.21 \\
    FedFTG & 65.31 & 88.36 & 71.85 & 90.39 & 65.10 & 86.47 & 71.82 & 90.62 & 46.87 & 60.53 & 53.15 & 63.43 \\
    FedPAC & 62.52 & 90.43 & 83.28 & 91.11 & 68.12 & 83.36 & 65.43 & 86.46 & 49.04 & 58.35 & 48.42 & 60.52 \\
    FedFM & 64.68 & 86.28 & 81.94 & 90.02 & 64.68 & 87.43 & 69.44 & 91.09 & 46.57 & 61.20 & 51.39 & 63.76 \\
    FedCCFA & 59.60 & 83.26 & 75.45 & 90.45 & 65.20 & 86.60 & 66.79 & 90.76 & 46.62 & 60.21 & 49.09 & 63.10 \\
    FedSA & 71.47 & 90.40 & 85.22 & 92.25 & 67.59 & 88.08 & 71.75 & 91.58 & 48.67 & 61.95 & 53.10 & 64.11 \\
    \midrule
    FedDF & 65.38 & 87.88 & 78.16 & 89.13 & 65.10 & 86.47 & 71.38 & 89.99 & 46.87 & 60.53 & 52.82 & 62.99 \\
    FedDC & 68.78 & 91.08 & 86.46 & 90.78 & 66.99 & 87.03 & 70.71 & 89.22 & 48.23 & 60.92 & 52.33 & 62.46 \\
    FedGen & 69.98 & 90.02 & 83.22 & 91.30 & 64.90 & 86.07 & 69.12 & 89.42 & 46.73 & 60.25 & 51.15 & 62.59 \\
    \midrule
    \textbf{FedTopo} & \textbf{74.96} & \textbf{91.30} & \textbf{89.88} & \textbf{94.46} & \textbf{73.87} & \textbf{92.76} & \textbf{77.89} & \textbf{96.63} & \textbf{53.20} & \textbf{65.93} & \textbf{57.64} & \textbf{67.64} \\
    \bottomrule
  \end{tabular}
  \caption{Test accuracy (\%) of global model under four types of non-IID partitioning. \textbf{N, L, K, Q} correspond to the following data partition types: (1) \textbf{N-skew (Amplitude Noise Skew)}: $\bar{\sigma} = 0.5$; (2) \textbf{L-skew (Label Skew)}: $\alpha_l = 0.1$; (3) \textbf{K-skew (Fixed-$k$ Label Skew)}: $k = 5$; (4) \textbf{Q-skew (Quantity Skew)}: $\alpha_q = 0.5$.}
  \label{tab:main_accuracy}
\end{table*}

\subsubsection{Visualization}

\begin{figure}[h!]
    \centering
    \includegraphics[width=0.48\textwidth]{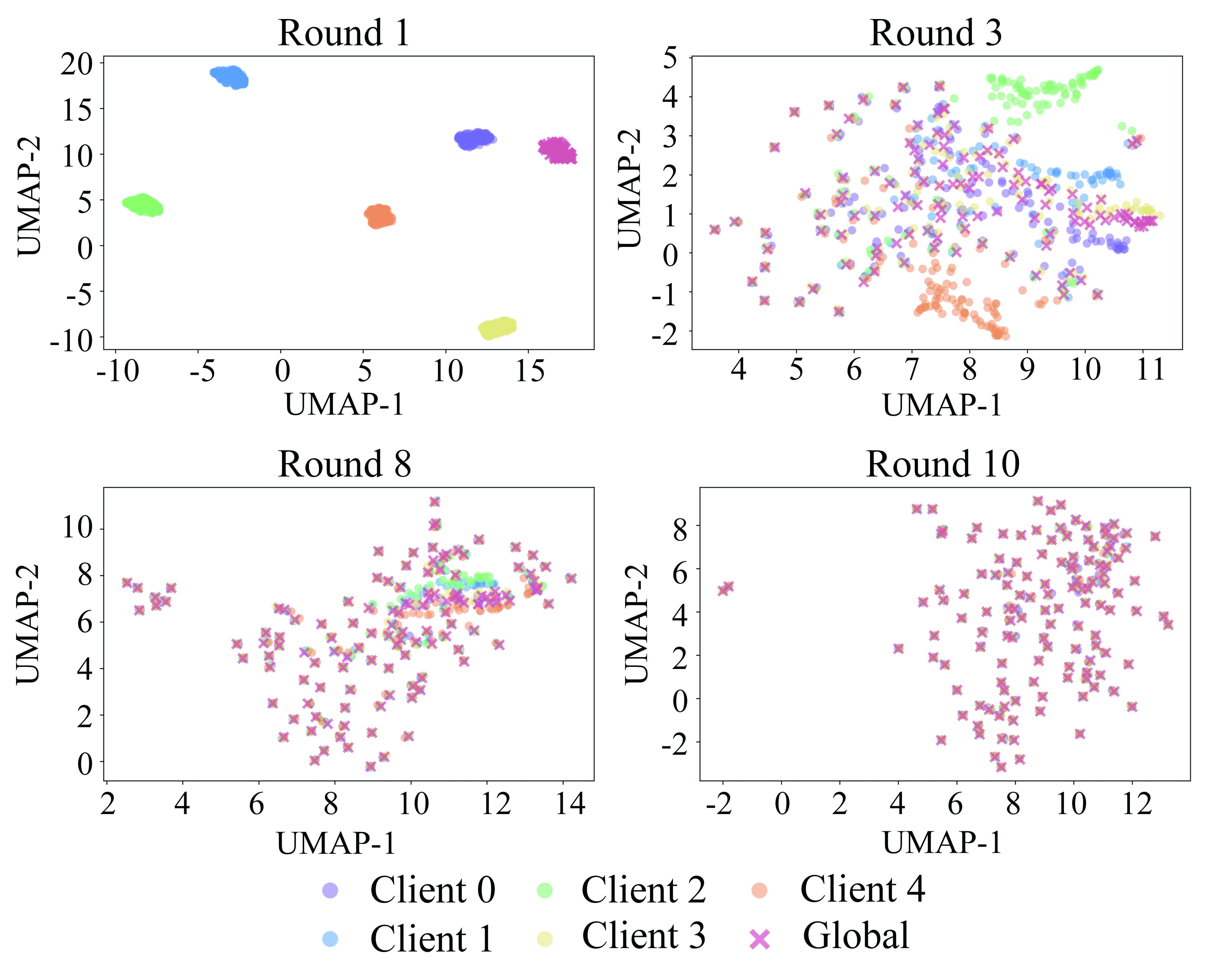}
    \caption{UMAP 2D feature projection of five clients and the global model at different communication rounds.}
    \label{fig:umap_proj}
\end{figure}

To verify the benefits of aligning local models with the global feature space, we visualize the global model’s representations using UMAP in Figure~\ref{fig:umap_proj}, with different colors for each client. At Round 0, client features are well separated and the global model lies far from them. By Round 2, client features begin to overlap with the global model, and by Round 9, the feature spaces nearly match, indicating successful alignment. We also visualize topological alignment using persistence barcodes in Appendix~\ref{appendix:additional-results}, which show increasing similarity between client and global topologies. These results confirm that TAL promotes topological consistency and mitigates representation drift.

\subsection{Ablation Study}

We evaluate the contribution of each core component in \textbf{FedTopo} through three ablation studies, focusing on \textbf{Topology-Guided Block Screening (TGBS)}, \textbf{Topological Regularization via TAL} and \textbf{Adaptive Scheduling of $\alpha$}.

\subsubsection{w/o TGBS: Effect of Block Selection}

To assess the effectiveness of TGBS, we evaluate FedTopo using features extracted from different candidate blocks: (1) ResNet-18: $\{\texttt{conv1}, \texttt{layer1}, \texttt{layer2}, \texttt{layer3}, \texttt{layer4}\}$; (2) CNN: $\{\texttt{conv1}, \texttt{conv2}\}$. Table~\ref{tab:tgbc_block_ablation} shows the following: (1) In ResNet-18, \texttt{layer2} performs best, followed by \texttt{layer3}. (2) In CNN, \texttt{conv1} outperforms \texttt{conv2}, indicating that overly deep layers may degrade topological signals in shallow networks. (3) AUC aligns well with accuracy, validating topological separability as a criterion for block selection. TGBS thus selects \texttt{layer2} for ResNet-18 and \texttt{conv1} for CNN, both achieving the best results, confirming its effectiveness.

\begin{table}[htbp]
  \centering
    \begin{tabular}{c|ccc}
    \toprule
    \textbf{Model} & \textbf{Blocks} & \textbf{AUC (\%)} & \textbf{Acc. (\%)} \\
    \midrule
    \multirow{5}{*}{ResNet-18} 
        & conv1   & 80.54 & 80.62 \\
        & layer1  & 81.29 & 82.13 \\
        & \textbf{layer2}  & \textbf{81.44} & \textbf{88.35} \\
        & layer3  & 80.13 & 86.76 \\
        & layer4  & 62.80 & 85.35 \\
    \midrule
    \multirow{2}{*}{CNN} 
        & \textbf{conv1}   & \textbf{61.41} & \textbf{87.52} \\
        & conv2   & 54.85 & 86.16 \\
    \bottomrule
    \end{tabular}
    \caption{Comparison of Topological Class Separability (AUC) and Federated Global Test Accuracy for Different Feature Blocks in ResNet-18 and CNN.}
  \label{tab:tgbc_block_ablation}
\end{table}

\begin{table}[htbp]
  \centering
  \begin{tabular}{cccc}
    \toprule
    \textbf{Methods}  & \textbf{Setting A} & \textbf{Setting B} & \textbf{Setting C} \\
    \midrule
    w/o TAL  & 53.90     & 58.87     & 47.23     \\
    \textbf{w/ TAL}   & \textbf{67.05}     & \textbf{73.67}     & \textbf{63.60}     \\
    \bottomrule
  \end{tabular}
  \caption{Test accuracy (\%) of FedTopo with and without TAL under three representative non-I.I.D.\ settings.}
  \label{tab:tal_ablation}
\end{table}

\begin{table}[htbp]
  \centering
  \setlength{\tabcolsep}{3.5pt} 
  \renewcommand{\arraystretch}{1.05} 
  \begin{tabular}{c|ccccc}
    \toprule
    \textbf{Setting} & \textbf{Round} & \textbf{Warm} & \textbf{P-wise} & \textbf{Linear} & \textbf{Smooth} \\
    \midrule
    \multirow{2}{*}{A}
      & 1   & 65.13 & 66.31 & \textbf{69.15} & 67.06 \\
      & 10  & 73.52 & 74.72 & 75.27 & \textbf{76.84} \\
    \midrule
    \multirow{2}{*}{B}
      & 1   & 58.32 & 59.48 & \textbf{62.60} & 60.86 \\
      & 10  & 67.21 & 68.62 & 69.94 & \textbf{71.31} \\
    \bottomrule
  \end{tabular}
  \caption{Performance (\%) of four $\alpha$ scheduling strategies on two CIFAR-10 non-I.I.D. partitions (Top-1 accuracy).}
  \label{tab:alpha_scheduling}
\end{table}

\subsubsection{w/o TAL: Effect of Topological Regularization} \label{sec:tal-ablation}

To evaluate the effect of Topological Alignment Loss (TAL), we compare FedTopo with and without TAL under three challenging non-I.I.D.\ scenarios: (A) ResNet-18 on CIFAR-10 with combined Q-skew ($\alpha_q = 0.5$) and K-skew ($k=2$), (B) Simple CNN on FMNIST with severe L-skew ($\alpha_l = 0.1$), and (C) ResNet-18 on CIFAR-100 with combined L-skew ($\alpha_l = 0.5$) and N-skew ($\sigma = 1.0$). The global model accuracies are reported in Table~\ref{tab:tal_ablation}. Across all settings, TAL consistently yields substantial accuracy gains, with an average improvement of $13.25\%$ (Setting~A: $+13.15\%$, Setting~B: $+14.80\%$, Setting~C: $+16.37\%$). These results demonstrate that TAL effectively reduces representation drift and enhances feature alignment in non-I.I.D.\  federated learning.

\subsubsection{Effect of Adaptive $\alpha$ Scheduling} \label{sec:alpha-study}

Table~\ref{tab:alpha_scheduling} compares four $\alpha$ scheduling strategies—\emph{Warm-up}, \emph{Piecewise}, \emph{Linear-Topo}, and \emph{Smooth-Topo}—under two non-I.I.D.\ settings on CIFAR-10 (Setting~A and Setting~B; see Section~\ref{sec:tal-ablation}).
We highlight three key observations. (1) In early rounds, \emph{Linear-Topo} yields the fastest performance gain, improving global accuracy by $+4\%$ over Warm-up in Round~1 of Setting~A (69.15\% vs.\ 65.13\%), demonstrating its effectiveness in quickly reducing cross-client representation gaps. (2) Over time, \emph{Smooth-Topo} consistently outperforms all other strategies, achieving the highest final accuracy in both settings (76.84\% / 71.31\%) and demonstrating greater stability with fewer fluctuations across rounds. (3) All topology-aware schedules outperform the naive Warm-up baseline, confirming the benefit of adaptive $\alpha$ scheduling in facilitating alignment throughout training.

\section{Conclusion}

This work introduces \textbf{FedTopo}, a topology-aware framework for federated learning under non-I.I.D.\ settings. By leveraging topological descriptors of intermediate feature maps, FedTopo aligns local representations with a global topological reference through \textbf{Topological Alignment Loss (TAL)}. To enable alignment, we propose \textbf{Topology-Guided Block Screening (TGBS)} for selecting the most discriminative layer, and construct a compact, differentiable \textbf{Topological Embedding (TE)} based on Persistence Images. Additionally, we incorporate an adaptive scheduling mechanism that dynamically adjusts regularization strength using topological feedback. Experiments on various datasets and partition schemes show that FedTopo effectively reduces representational drift and improves generalization. Theoretically, we establish the stability of TE and show that TAL preserves convergence under mild assumptions. Overall, our results highlight the potential of topological signals as a principled regularization for robust federated learning. Future work will extend FedTopo to graphs and sequences, develop privacy-preserving TE aggregation with online TGBS for large-scale deployments, and explore applications in decentralized data ecosystems, enabling secure data ownership verification, trading, and value extraction from multi-party data while preserving privacy.

\section*{Acknowledgments}

This work was supported by the Shanghai Science and Technology Innovation Action Plan (24BC3200600) and National Natural Science Foundation of China (62441227).

\bibliography{aaai2026}



\onecolumn

\appendix
\section{Proofs of Theoretical Results}
\label{appendix:theory}

\renewcommand{\theequation}{A.\arabic{equation}}

\subsection{AUC/ROC Superiority Implies a Monotonic Increase in a Mutual Information Lower Bound}

We formalize the connection between topological class separability—quantified by AUC/ROC—and information content. Throughout, let $Y \in \{0,1\}$ denote the class label with prior $\pi^+ = \Pr(Y = 1)$ and $\pi^- = \Pr(Y = 0) = 1 - \pi^+$. Let $T_b = T_b(X)$ denote the representation at block $b$, and let $s_b : \mathcal{T} \to \mathbb{R}$ be a scalar score used to construct the ROC curve and compute AUC. Denote the entropy of $Y$ by $H(Y)$ and the binary entropy function by $h_b(\cdot)$.

\paragraph{Assumptions.}
\begin{enumerate}[label=(A\arabic*)]
    \item \textbf{(Neyman--Pearson compatibility):} The score $s_b(T_b)$ is a strictly monotonic function of the log-likelihood ratio, i.e., $s_b(T_b) \propto \log \frac{p(T_b \mid Y = 1)}{p(T_b \mid Y = 0)}$. This ensures that thresholding $s_b$ yields the uniformly most powerful test.
    \item \textbf{(ROC dominance):} For two blocks $b_1$ and $b_2$, the ROC curve of $s_{b_1}(T_{b_1})$ dominates that of $s_{b_2}(T_{b_2})$ pointwise, with strict inequality over a subset of false positive rates of nonzero measure. As a consequence, $\mathrm{AUC}_{b_1} > \mathrm{AUC}_{b_2}$.
\end{enumerate}

\begin{proposition}[Monotonicity of MI Lower Bound via Bayes Error and Fano’s Inequality]
\label{prop:roc-mi}
Under Assumptions \textnormal{(A1)} and \textnormal{(A2)}, let $e_b^\star$ denote the Bayes error corresponding to the optimal threshold on $s_b(T_b)$. Then:
\[
e_{b_1}^\star < e_{b_2}^\star.
\]
Moreover, mutual information admits the standard lower bound:
\[
I(T_b; Y) \ge H(Y) - h_b(e_b^\star),
\]
which implies:
\[
H(Y) - h_b(e_{b_1}^\star) > H(Y) - h_b(e_{b_2}^\star).
\]
Hence, the lower bound on $I(T_b; Y)$ is strictly larger for $b_1$ than for $b_2$.
\end{proposition}

\paragraph{Proof.}
By Assumption \textnormal{(A1)}, thresholding $s_b$ produces the Neyman--Pearson optimal test. The ROC curve for $s_b$ is thus the best achievable for any test based on $T_b$. Assumption \textnormal{(A2)} ensures that the ROC curve of $b_1$ dominates that of $b_2$, with strict dominance on a set of FPRs of positive measure. This implies not only a larger AUC but also a strictly smaller Bayes error $e^\star$ for $b_1$, following classical decision theory.

Fano’s inequality gives a lower bound on mutual information:
\[
I(T_b; Y) \ge H(Y) - h_b(e_b),
\]
for any decoder with average error $e_b$. Applying this to the optimal decoder yields:
\[
I(T_b; Y) \ge H(Y) - h_b(e_b^\star).
\]
Since $h_b(\cdot)$ is strictly increasing on $[0, \tfrac{1}{2}]$, it follows that $e_{b_1}^\star < e_{b_2}^\star$ implies:
\[
H(Y) - h_b(e_{b_1}^\star) > H(Y) - h_b(e_{b_2}^\star).
\]

\begin{corollary}[Equal Class Priors and Total Variation Distance]
If $\pi^+ = \pi^- = \tfrac{1}{2}$, then the Bayes error is $e_b^\star = \tfrac{1}{2}(1 - \mathrm{TV}(P^+, P^-))$, where $\mathrm{TV}$ denotes the total variation distance. Proposition~\ref{prop:roc-mi} then implies:
\[
I(T_{b_1}; Y) \ge \log 2 - h_b(e_{b_1}^\star) > \log 2 - h_b(e_{b_2}^\star).
\]
\end{corollary}

\begin{remark}[Applicability to Distance-Based AUCs]
In practice, AUC is often computed from pairwise distances (e.g., between same-class vs.\ different-class pairs). Proposition~\ref{prop:roc-mi} applies if the induced scalar score is strictly monotonic in the log-likelihood ratio. This holds, for instance, in homoscedastic Gaussian models (under Mahalanobis distance) or when a projection yields Fisher-optimal separation. If this monotonicity fails, AUC computed from distances may not reflect the Neyman--Pearson envelope, and the proposition does not hold.
\end{remark}

\noindent\textbf{Takeaway.} Under mild and interpretable assumptions—specifically, Neyman--Pearson compatibility and strict ROC dominance—higher AUC implies strictly lower Bayes error and a strictly larger lower bound on $I(T_b; Y)$. Therefore, maximizing AUC serves as a valid and principled proxy for maximizing mutual information.

\subsection{Proof of Lipschitz Stability of Topological Embedding}

We formalize that the \emph{topological embedding} (TE), constructed from channel-wise persistence images (PIs), is Lipschitz continuous with respect to the input. We state the result for the channel-averaged TE used in the main text; the concatenated variant follows analogously.

\begin{proposition}[Lipschitz stability of TE]\label{prop:lipschitz-te}
Let $\phi_\ell(x;\mathbf{w})\in\mathbb{R}^{C\times H\times W}$ be the activation at layer $\ell$ of a CNN with fixed parameters $\mathbf{w}$, and let $\phi_\ell^{(j)}(x)\in\mathbb{R}^{H\times W}$ denote its $j$-th channel. For each channel, let $\mathrm{PI}\!\left(\phi_\ell^{(j)}(x)\right)\in\mathbb{R}^M$ be the persistence image (PI) computed from the sublevel-set cubical filtration with a fixed Gaussian kernel and a fixed rasterization grid.

Define the \emph{channel-averaged} TE by
\[
\mathbf{t}_{\mathrm{avg}}(x)
\;:=\;
\frac{1}{C}\sum_{j=1}^C \mathrm{PI}\!\left(\phi_\ell^{(j)}(x)\right)
\;\in\;\mathbb{R}^M.
\]
Assume standard TDA stability conditions (e.g., $\,\phi_\ell^{(j)}(x)$ are tame so that persistence diagrams are well-defined and stable), and assume the network mapping $x\mapsto \phi_\ell(x)$ is Lipschitz:
\[
\|\phi_\ell(x)-\phi_\ell(x')\|_2 \;\le\; L_\phi\,\|x-x'\|_2.
\]
Then there exists a constant $L_{\mathrm{PI}}>0$ (depending only on the PI kernel and grid) such that
\begin{equation}
\big\|\mathbf{t}_{\mathrm{avg}}(x)-\mathbf{t}_{\mathrm{avg}}(x')\big\|_2
\;\le\;
L_{\mathrm{PI}}\,L_\phi\,\|x-x'\|_2.
\label{eq:te-lipschitz}
\end{equation}
In particular, the TE is Lipschitz with respect to the input.
\end{proposition}

\paragraph{Proof.}
\textbf{Step 1: PI stability for a single channel.}
Let $f,g:\Omega\to\mathbb{R}$ be two scalar fields on the $H\times W$ grid (e.g., $f=\phi_\ell^{(j)}(x)$ and $g=\phi_\ell^{(j)}(x')$). By the stability of persistence diagrams for sublevel-set filtrations (e.g., Cohen--Steiner et al.), the bottleneck distance between the diagrams satisfies
\[
d_B\!\big(D(f),D(g)\big)\;\le\;\|f-g\|_\infty.
\]
Moreover, the PI map (with fixed Gaussian kernel and grid) is Lipschitz with respect to the diagram distance (e.g., Adams et al.), hence there exists $L_{\mathrm{PI}}>0$ such that
\[
\big\|\mathrm{PI}(f)-\mathrm{PI}(g)\big\|_2 \;\le\; L_{\mathrm{PI}}\,d_B\!\big(D(f),D(g)\big)
\;\le\; L_{\mathrm{PI}}\,\|f-g\|_\infty.
\]
Applying this with $f=\phi_\ell^{(j)}(x)$ and $g=\phi_\ell^{(j)}(x')$ yields
\begin{equation}
\big\|\mathrm{PI}\!\left(\phi_\ell^{(j)}(x)\right)-\mathrm{PI}\!\left(\phi_\ell^{(j)}(x')\right)\big\|_2
\;\le\;
L_{\mathrm{PI}}\;\big\|\phi_\ell^{(j)}(x)-\phi_\ell^{(j)}(x')\big\|_\infty.
\label{eq:pi-chan}
\end{equation}

\textbf{Step 2: Lipschitz continuity of the feature map.}
By composition of Lipschitz layers (convolutions/linear maps with bounded operator norms and 1-Lipschitz activations such as ReLU), there exists $L_\phi$ such that
\[
\|\phi_\ell(x)-\phi_\ell(x')\|_2 \;\le\; L_\phi\,\|x-x'\|_2.
\]
Since the $j$-th channel is a sub-tensor of $\phi_\ell(x)$, after flattening we have the norm relation
\[
\big\|\phi_\ell^{(j)}(x)-\phi_\ell^{(j)}(x')\big\|_\infty
\;\le\;
\big\|\phi_\ell(x)-\phi_\ell(x')\big\|_2
\;\le\;
L_\phi\,\|x-x'\|_2.
\]
Combining with \eqref{eq:pi-chan} gives, for every $j$,
\begin{equation}
\big\|\mathrm{PI}\!\left(\phi_\ell^{(j)}(x)\right)-\mathrm{PI}\!\left(\phi_\ell^{(j)}(x')\right)\big\|_2
\;\le\; L_{\mathrm{PI}}\,L_\phi\,\|x-x'\|_2.
\label{eq:pi-chan2}
\end{equation}

\textbf{Step 3: From channels to the averaged TE.}
By the triangle inequality,
\[
\big\|\mathbf{t}_{\mathrm{avg}}(x)-\mathbf{t}_{\mathrm{avg}}(x')\big\|_2
\;=\;
\Big\|\frac{1}{C}\sum_{j=1}^C
\big(\mathrm{PI}\!\left(\phi_\ell^{(j)}(x)\right)-\mathrm{PI}\!\left(\phi_\ell^{(j)}(x')\right)\big)
\Big\|_2
\;\le\;
\frac{1}{C}\sum_{j=1}^C
\big\|\mathrm{PI}\!\left(\phi_\ell^{(j)}(x)\right)-\mathrm{PI}\!\left(\phi_\ell^{(j)}(x')\right)\big\|_2.
\]
Using \eqref{eq:pi-chan2} and canceling the factor $1/C$ yields
\[
\big\|\mathbf{t}_{\mathrm{avg}}(x)-\mathbf{t}_{\mathrm{avg}}(x')\big\|_2
\;\le\;
L_{\mathrm{PI}}\,L_\phi\,\|x-x'\|_2,
\]
which is precisely \eqref{eq:te-lipschitz}.

\begin{remark}[Concatenated TE]\label{rem:concat}
If one defines the TE by concatenation
\[
\mathbf{t}_{\mathrm{cat}}(x)
:=
\big[
\mathrm{PI}\!\left(\phi_\ell^{(1)}(x)\right),\ldots,
\mathrm{PI}\!\left(\phi_\ell^{(C)}(x)\right)
\big]\in\mathbb{R}^{MC},
\]
then, stacking the channel-wise bounds \eqref{eq:pi-chan2} and using $\|\cdot\|_\infty\le\|\cdot\|_2$ on the flattened activation tensor, one obtains
\[
\big\|\mathbf{t}_{\mathrm{cat}}(x)-\mathbf{t}_{\mathrm{cat}}(x')\big\|_2
\;\le\;
L_{\mathrm{PI}}\,\big\|\phi_\ell(x)-\phi_\ell(x')\big\|_2
\;\le\;
L_{\mathrm{PI}}\,L_\phi\,\|x-x'\|_2.
\]
Hence the concatenated TE is also Lipschitz with the same order constant.
\end{remark}

\noindent\textbf{Discussion.} The constant $L_{\mathrm{PI}}$ depends only on the PI construction (kernel bandwidth and grid resolution), while $L_\phi$ depends on the product of operator norms of the layers up to $\ell$. The proof relies on the standard stability chain: function perturbation $\Rightarrow$ diagram stability (bottleneck) $\Rightarrow$ PI stability, composed with the Lipschitz continuity of the CNN feature map.

\subsection{Proof of FedProx-style Convergence under TAL}

We show that the proposed \emph{Topological Alignment Loss (TAL)} acts as a proximal regularizer and thus admits a FedProx-style convergence guarantee under standard smoothness and local strong convexity assumptions.

\begin{proposition}[FedProx-style convergence with TAL]\label{prop:fedprox_tal}
Assume for each client $i$:
\begin{itemize}
    \item $F_i(w)$ is $L$-smooth, i.e., $\|\nabla F_i(w)-\nabla F_i(w')\|\le L\|w-w'\|$ for all $w,w'$.
    \item The TAL term is defined as an expectation over local data:
    \[
    \mathcal{L}_{\mathrm{topo}}(w;\tilde w)\;:=\;\mathbb{E}_{x\sim\mathcal{D}_i}\!\left[\;\big\|\mathbf{t}(x;w)-\mathbf{t}(x;\tilde w)\big\|_2^2\right],
    \]
    where $\tilde w$ is a reference (e.g., last global model).
    \item (\textbf{Local strong convexity}) There exists a neighborhood $\mathcal{N}(\tilde w)$ and $\mu>0$ such that $\mathcal{L}_{\mathrm{topo}}(\cdot;\tilde w)$ is $\mu$-strongly convex on $\mathcal{N}(\tilde w)$.
    \item (\textbf{Smoothness of TAL}) $\mathcal{L}_{\mathrm{topo}}(\cdot;\tilde w)$ is $\beta$-smooth on $\mathcal{N}(\tilde w)$, i.e., its gradient is $\beta$-Lipschitz with respect to $w$ in that neighborhood.\footnote{This holds, e.g., if the TE map $w\mapsto \mathbf{t}(x;w)$ is $C^1$ with locally Lipschitz Jacobian and uniformly bounded first/second derivatives, and the data expectation is well-defined.}
\end{itemize}
Consider the locally regularized objective
\[
F_i^{\mathrm{reg}}(w)\;:=\;F_i(w)\;+\;\alpha\,\mathcal{L}_{\mathrm{topo}}(w;\tilde w),
\]
with $\alpha>0$. Then on $\mathcal{N}(\tilde w)$, $F_i^{\mathrm{reg}}$ is $(\alpha\mu)$-strongly convex and $(L+\alpha\beta)$-smooth. Consequently, gradient descent with a fixed step size $\eta\le 1/(L+\alpha\beta)$ satisfies the linear rate
\[
F_i^{\mathrm{reg}}(w^{(t)})-F_i^{\mathrm{reg}}(w^\star)\;\le\;\Big(1-\eta\,\alpha\mu\Big)^t\;\big(F_i^{\mathrm{reg}}(w^{(0)})-F_i^{\mathrm{reg}}(w^\star)\big),
\]
and the iteration complexity to reach an $\varepsilon$-accurate solution is
\begin{equation}
t\;=\;O\!\left(\frac{L+\alpha\beta}{\alpha\mu}\,\log\frac{1}{\varepsilon}\right).
\label{eq:tal-lin-rate}
\end{equation}
Moreover, in the federated setting, if in each communication round every client approximately solves its local subproblem $\min_w F_i^{\mathrm{reg}}(w)$ within a fixed accuracy and the server aggregates the updated models (e.g., weighted averaging as in FedAvg/FedProx), then the global model enjoys a FedProx-style convergence with the same order of complexity as in \eqref{eq:tal-lin-rate}, replacing the proximal coefficient by $\alpha\mu$ and the smoothness constant by $L+\alpha\beta$.
\end{proposition}

\paragraph{Proof.}
By assumptions, on $\mathcal{N}(\tilde w)$ the TAL term is $\mu$-strongly convex and $\beta$-smooth. Hence $F_i^{\mathrm{reg}}(w)=F_i(w)+\alpha\,\mathcal{L}_{\mathrm{topo}}(w;\tilde w)$ is $(\alpha\mu)$-strongly convex and $(L+\alpha\beta)$-smooth on $\mathcal{N}(\tilde w)$. Standard results for smooth and strongly convex functions (e.g., Nesterov, 2004) imply that gradient descent with $\eta\le 1/(L+\alpha\beta)$ yields
\[
F_i^{\mathrm{reg}}(w^{(t)})-F_i^{\mathrm{reg}}(w^\star)\;\le\;\Big(1-\eta\,\alpha\mu\Big)^t\;\big(F_i^{\mathrm{reg}}(w^{(0)})-F_i^{\mathrm{reg}}(w^\star)\big),
\]
which gives the iteration complexity \eqref{eq:tal-lin-rate}.

For the federated convergence, we follow the FedProx argument: if each client (approximately) minimizes its locally strongly convex and smooth subproblem $F_i^{\mathrm{reg}}$ to a fixed accuracy per round and the server aggregates the local solutions, then the global objective decreases geometrically with a rate governed by the condition number $(L+\alpha\beta)/(\alpha\mu)$ (cf.\ Li et al.\ 2020). The only change is that the classical quadratic proximal term $\tfrac{\mu}{2}\|w-\tilde w\|^2$ is replaced by $\alpha\,\mathcal{L}_{\mathrm{topo}}(w;\tilde w)$, which acts as a proximal potential thanks to the local $\mu$-strong convexity; thus the same proof strategy applies with constants $\mu'\!=\alpha\mu$ and $L'\!=L+\alpha\beta$.

\paragraph{Remarks.}
(i) The local strong convexity of $\mathcal{L}_{\mathrm{topo}}(\cdot;\tilde w)$ can be ensured, for instance, if the TE map $w\mapsto \mathbf{t}(x;w)$ is locally bi-Lipschitz and its Jacobian has a uniformly bounded-away-from-zero smallest singular value near $\tilde w$ (after data averaging), so that the Hessian of $w\mapsto \tfrac{1}{2}\|\mathbf{t}(x;w)-\mathbf{t}(x;\tilde w)\|^2$ dominates a positive multiple of the identity.  
(ii) If one absorbs $\alpha\beta$ into the symbol $L$ (redefining $L\leftarrow L+\alpha\beta$), the complexity can be written compactly as $O\!\big(\tfrac{L}{\alpha\mu}\log\tfrac{1}{\varepsilon}\big)$; we keep both terms explicit to highlight the dependence on $\alpha$.

\section{Details of Proposed Method} \label{appendix:method-details}

\setcounter{figure}{0}
\setcounter{table}{0}
\setcounter{equation}{0}
\renewcommand{\thefigure}{B.\arabic{figure}}
\renewcommand{\thetable}{B.\arabic{table}}
\renewcommand{\theequation}{B.\arabic{equation}}

\subsection{Topology-Guided Block Screening (TGBS)} \label{appendix:tgbc}

\subsubsection{Motivation}
We hypothesize that if a layer’s topological signature consistently separates within-class from between-class pairs, the layer encodes task-relevant semantics. FedTopo therefore prioritizes alignment at such a \emph{topology-informative} block, expecting more beneficial regularized gradients. To this end, we introduce \textbf{Topology-Guided Block Screening (TGBS)}, which automatically selects the block with the highest \emph{topological class discriminability}. This leverages the well-established link between persistent-homology signatures and class separability: persistent features capture global structures (connectivity, holes) that are robust to deformations and noise and complement pixel-/patch-level objectives \shortcite{carriere2020perslay, moor2023topological, mosinska2018beyond, clough2019topological}.

\subsubsection{TGBS Design}
On a small validation subset of CIFAR-10, we evaluate each candidate block (\texttt{conv1} through \texttt{layer4}) of ResNet-18. For each block, we randomly sample within-class and between-class pairs, project activations (for channel compression), compute cubical persistent homology, and compare topologies using three families of metrics: bottleneck distance \shortcite{cohen2007stability}, Wasserstein distance \shortcite{carriere2017sliced}, and PI embeddings with Euclidean/cosine distances \shortcite{adams2017persistence}. We set similarity as $s=-d$ and score discriminability by ROC-AUC.

\begin{figure}[htbp]
    \centering
    \begin{minipage}[t]{0.48\linewidth}
        \centering
        \includegraphics[width=\linewidth]{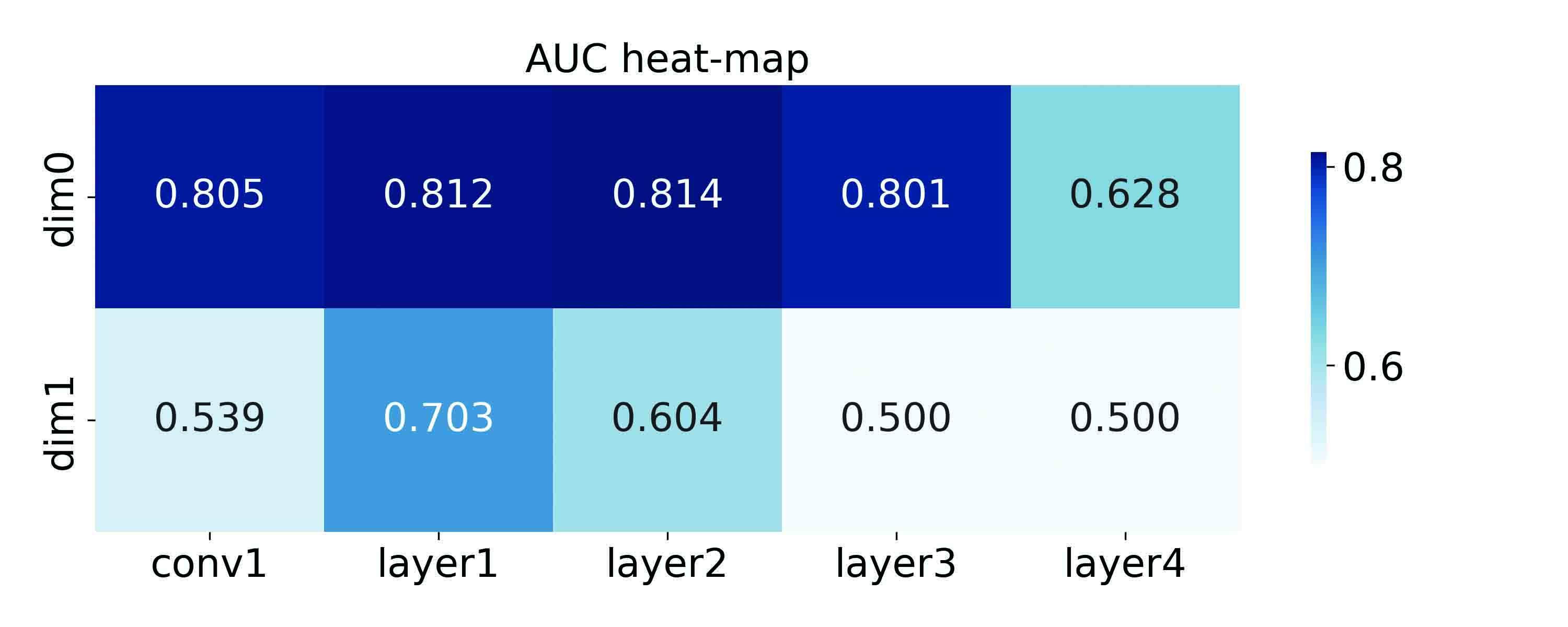}
        \caption{AUC heatmap across backbone layers and reduced dimensions. The best AUC (\textbf{0.814}) occurs at \texttt{layer2} (dim 0), indicating the most topology-informative block for class separation.}
        \label{fig:heatmap_auc}
    \end{minipage}\hfill
    \begin{minipage}[t]{0.48\linewidth}
        \centering
        \includegraphics[width=\linewidth]{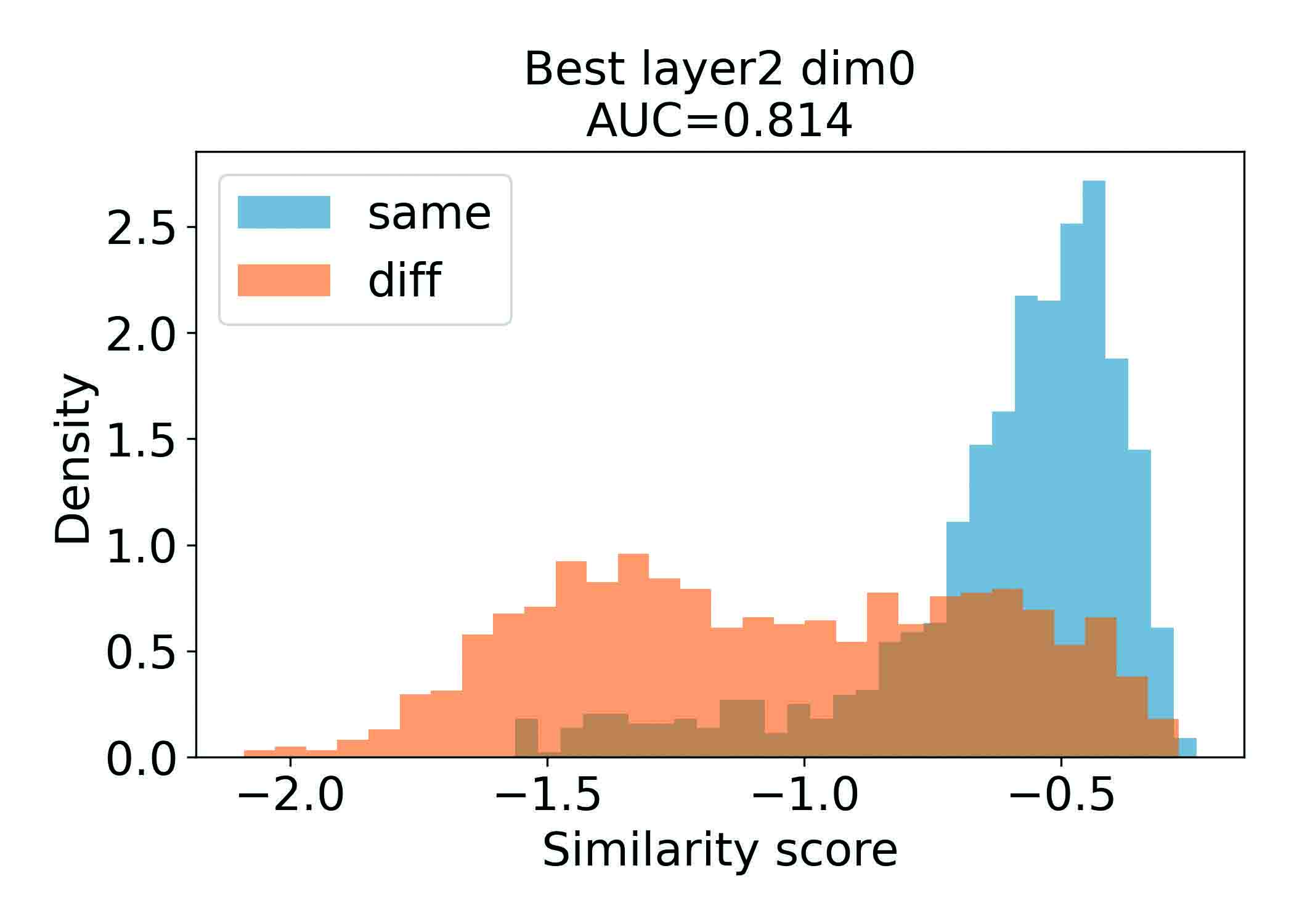}
        \caption{Same-class vs.\ different-class similarity distributions at \texttt{layer2} (dim 0). The area under the ROC curve is $0.814$.}
        \label{fig:hist_layer2_dim0}
    \end{minipage}
\end{figure}

Figure~\ref{fig:heatmap_auc} summarizes the AUC across blocks and reduced dimensions, with \texttt{layer2} attaining the highest AUC (\(0.814\)). Shallower blocks show weaker topology due to local texture noise; the top block (\texttt{layer4}) slightly degrades under strong channel compression, suggesting that over-smoothed semantics may attenuate persistent patterns. We thus select \texttt{layer2} for feature extraction in FedTopo. Figure~\ref{fig:hist_layer2_dim0} further shows clear separation between within- and between-class distance distributions for the most discriminative dimension.

For other backbones we apply the same screening protocol; selected blocks are reported in Appendix Subsection~\ref{appendix:feature-block-selection}.

\paragraph{Topology distance metrics.}
To evaluate class-discriminative power within TGBS (Section~\ref{sec:tgbsmethod}), we compare persistence diagrams using:
\[
d_B(\mathcal{D}_1, \mathcal{D}_2) \;=\; \inf_{\gamma} \sup_{x \in \mathcal{D}_1} \|x - \gamma(x)\|_\infty,\qquad
d_p(\mathcal{D}_1, \mathcal{D}_2) \;=\; \Big(\inf_{\gamma} \sum_{x \in \mathcal{D}_1} \|x - \gamma(x)\|^{p}\Big)^{1/p}.
\]
These metrics quantify geometric discrepancies between topological summaries and are used to compute ROC-AUC for layer selection.

\subsection{Adaptive $\alpha$ Scheduling Strategies}
\label{app:alpha-scheduling}
To balance the cross-entropy loss $\mathcal{L}_{\mathrm{CE}}$ and the topological alignment loss $\mathcal{L}_{\mathrm{topo}}$, we define the per-client objective at communication round $r$ and local epoch $e$ as
\begin{equation}
\mathcal{L}_i^{(r,e)} \;=\; \mathcal{L}_{\mathrm{CE}} \;+\; \alpha_i^{(r,e)}\,\mathcal{L}_{\mathrm{topo}},
\label{eq:B.alpha.obj}
\end{equation}
where $\alpha_i^{(r,e)}$ controls the strength of topological regularization. We compute $\alpha_i^{(r,e)}$ dynamically using both client-specific statistics and historical optimization signals, combining warm-up, round decay, and strategy-specific modulation.

Let $\alpha_{\max}$ be a global upper bound; let $\alpha_{\min}^{(i)}$ be a client-specific lower bound (from label imbalance); let $E_{\text{warm}}$ be the warm-up epochs; and let $\gamma$ be the round-wise decay rate. Define
\begin{equation}
\lambda^{(r,e)} \;=\; \min\!\Big(1, \tfrac{e+1}{E_{\text{warm}}}\Big)\cdot e^{-\gamma r}, \qquad
\alpha_i^{(r,e)} \;=\; \lambda^{(r,e)}\,\alpha_i^{\mathrm{base}}.
\label{eq:B.alpha.lambda}
\end{equation}
We instantiate $\alpha_i^{\mathrm{base}}$ via four strategies:

\paragraph{Strategy 1: Warm-up.}
\(\alpha_i^{\mathrm{base}}=\alpha_{\max}\). No topological feedback; only warm-up and decay schedule.

\paragraph{Strategy 2: Linear-Topo.}
Let $\bar{\ell}_{\text{topo}}^{(i)}$ be the smoothed TAL over the last $w$ local steps:
\[
\bar{\ell}_{\text{topo}}^{(i)} \;=\; \tfrac{1}{w}\sum_{t=e-w+1}^{e}\mathcal{L}_{\text{topo}}^{(i,t)},\qquad
\eta_i \;=\; \Big(\tfrac{\bar{\ell}_{\text{topo}}^{(i)} - \ell_{\min}}{\ell_{\max}-\ell_{\min}+\epsilon}\Big)^{\beta},~~\beta\in(0,1],
\]
\[
\alpha_i^{\mathrm{base}} \;=\; \alpha_{\min}^{(i)} + \big(\alpha_{\max}-\alpha_{\min}^{(i)}\big)\,\eta_i.
\]

\paragraph{Strategy 3: Piecewise.}
With the same $\bar{\ell}_{\text{topo}}^{(i)}$,
\[
\alpha_i^{\mathrm{base}}=
\begin{cases}
\alpha_{\min}^{(i)}, & \bar{\ell}_{\text{topo}}^{(i)} \le \ell_{\min},\\
\alpha_{\max}, & \bar{\ell}_{\text{topo}}^{(i)} \ge \ell_{\max},\\
\alpha_{\min}^{(i)}+\dfrac{\bar{\ell}_{\text{topo}}^{(i)}-\ell_{\min}}{\ell_{\max}-\ell_{\min}+\epsilon}\,\big(\alpha_{\max}-\alpha_{\min}^{(i)}\big), & \text{otherwise.}
\end{cases}
\]

\paragraph{Strategy 4: Smooth-Topo.}
Use a longer EWMA window to stabilize the signal:
\[
\eta_i \;=\; \Big(\tfrac{\tilde{\ell}_{\text{topo}}^{(i)} - \ell_{\min}}{\ell_{\max}-\ell_{\min}+\epsilon}\Big)^{\beta},\quad
\alpha_i^{\mathrm{base}} \;=\; \alpha_{\min}^{(i)} + \big(\alpha_{\max}-\alpha_{\min}^{(i)}\big)\,\eta_i,
\]
where $\tilde{\ell}_{\text{topo}}^{(i)}$ is the EWMA of recent TAL values. Smoothing improves robustness to transient spikes.

\subsection{Computation and Communication Cost}\label{appendix:cost}
We summarize worst-case and practical costs for 2-D cubical complexes and the resulting communication overhead.

\paragraph{Per-block persistent homology.}
For spatial size $H\times W$ and $C$ channels, if the scalar field is defined on the $H\times W$ grid vertices, the per-channel cubical complex contains
\begin{equation}
m_{\text{cell}}
= \underbrace{HW}_{\text{vertices}}
+ \underbrace{[H(W-1)+W(H-1)]}_{\text{edges}}
+ \underbrace{(H-1)(W-1)}_{\text{squares}}
= 4HW - 2H - 2W + 1,
\label{eq:B.cell-count}
\end{equation}
hence $m=\Theta(HW)$.

\begin{itemize}
\item \textbf{Worst-case time.} Boundary-matrix reduction yields
\begin{equation}
T_{\text{PH}}^{\text{worst}}=O\!\big(C\,m^{\omega}\big)=O\!\big(C\,(HW)^{\omega}\big),\quad \omega\le 3
\label{eq:B.ph-worst}
\end{equation}
(e.g., with clearing/compression heuristics typically $\omega\approx 2.3$).

\item \textbf{Practical 2-D cost.} With union-find for $H_0$ and clearing/Morse heuristics for $H_1$, the observed cost is near-linear:
\begin{equation}
T_{\text{PH}}^{\text{prac}}=\tilde O(CHW),
\label{eq:B.ph-prac}
\end{equation}
where $\tilde O(\cdot)$ hides polylog factors (e.g., threshold sorting).
\end{itemize}

\paragraph{PI and TE construction.}
Given a diagram for channel $c$, forming a PI $\psi_c\in\mathbb{R}^{M}$ costs $O(|\mathrm{PD}_c|+M)$; averaging over $K$ kept channels to obtain the TE (Eq.~\eqref{eq:topo-embedding}) costs $O(KM)$.

\paragraph{Numerical scale.}
For $H=W=8$ (ResNet-18 \texttt{layer2} output after downsampling) and $C=256$, Eq.~\eqref{eq:B.cell-count} gives $m_{\text{cell}}=225$ per channel; the block totals $\approx 225\times 256=5.76\times 10^4$ cells. The extra time in practice is dominated by one additional forward pass of the selected block to compute $\mathbf{t}_i(x;\bar{\mathbf{w}})$.

\paragraph{Communication overhead.}
FedTopo does \emph{not} transmit per-sample topological data: both $\mathbf{t}_i(x;\mathbf{w}_i)$ and $\mathbf{t}_i(x;\bar{\mathbf{w}})$ are computed locally. If one TE (length $M$) is uploaded per round, the uplink fraction relative to model upload is
\begin{equation}
\Delta_{\text{FedTopo}}=\frac{M}{|\mathbf{w}|}.
\label{eq:B.comm-ratio}
\end{equation}
With $M=64$:
\begin{itemize}
\item ResNet-18 (CIFAR variant, $|\mathbf{w}|\approx 11$M): $\Delta\approx 5.8\times 10^{-6}$ ($0.00058\%$).
\item SimpleCNN for MNIST ($|\mathbf{w}|\approx 0.1$M): $\Delta\approx 6.4\times 10^{-4}$ ($0.064\%$).
\end{itemize}
Thus the communication cost is $10^{-6}$–$10^{-4}$ of model upload per round and scales linearly with the number of uploaded TE summaries.

\paragraph{Takeaway.}
The PH step admits a worst-case bound $O\!\big(C(HW)^{\omega}\big)$ yet behaves near-linearly $\tilde O(CHW)$ in 2-D practice. The communication overhead is negligible (sub-$0.1\%$ even for small models). FedTopo mainly adds one extra forward of the selected block plus PI/TE construction, leaving the communication pattern of FedAvg unchanged.

\section{Experimental Details} \label{appendix:experimental-details}

\setcounter{figure}{0}
\setcounter{table}{0}
\setcounter{equation}{0}
\renewcommand{\thefigure}{C.\arabic{figure}}
\renewcommand{\thetable}{C.\arabic{table}}
\renewcommand{\theequation}{C.\arabic{equation}}

\subsection{Data Partitions} \label{appendix:data-partitions}

\paragraph{Data-partition schemes}

We evaluate our method under three canonical non-I.I.D.\ partitions. We assume the following common notations across all partitions:

\begin{itemize}
    \item $N$: Total number of samples in the dataset.
    \item $K$: Total number of classes.
    \item $C = \{0, 1, \dots, n-1\}$: Set of $n$ clients.
    \item $\mathcal{D} = \{(x_i, y_i)\}_{i=1}^N$: Full dataset, where $x_i$ are the features and $y_i$ are the labels.
    \item $\mathcal{I}_c = \{ i \mid y_i = c \}$: Index set for samples of class $c$.
\end{itemize}

The non-I.I.D.\ partitions considered are as follows:

\textbf{Q-skew (quantity skew).} Draw
\[
\mathbf{p} = (p_1, \dots, p_K) \sim \operatorname{Dir}_K(\alpha_q), \qquad
\hat{p}_j = \frac{p_j}{\sum_{m=1}^{K} p_m}.
\]
With total sample size $N$, client $C_j$ receives
\[
I_j = \Bigl[\sum_{m<j}\hat{p}_m N,\; \sum_{m\le j}\hat{p}_m N\Bigr), \quad
\mathcal{D}_j = \{ (x_i, y_i) \mid i \in I_j \}.
\]

\textbf{L-skew (label skew).} For each class $c \in [C]$ sample
\[
\mathbf{p}_c = (p_{c,1}, \dots, p_{c,K}) \sim \operatorname{Dir}_K(\alpha_\ell), \qquad
\hat{p}_{c,j} = \frac{p_{c,j}}{\sum_m p_{c,m}}.
\]
Let $N_c$ be the number of examples in class $c$; assign to client $C_j$
\[
I_{c,j} = \Bigl[\sum_{m<j}\hat{p}_{c,m} N_c,\; \sum_{m\le j}\hat{p}_{c,m} N_c\Bigr)
\]
\[
\mathcal{D}_{c,j} = \{ (x_i, y_i) \mid i \in I_{c,j} \}
\]
and set $\mathcal{D}_j = \bigcup_c \mathcal{D}_{c,j}$.

\textbf{N-skew (amplitude noise).} Let $K$ be the number of clients, $\bar{\sigma} > 0$ the maximum noise scale, and $x_i \in \mathbb{R}^{d}$ an input. Client $C_j$ holds
\[
\mathcal{D}_j = \bigl\{ (x_i + n_{i,j},\; y_i) \mid i \in I_j \bigr\}, \qquad
n_{i,j} \sim \mathcal{N}\!\bigl(0, \sigma_j^{2} I_d \bigr),
\]
with
\[
\sigma_j = \frac{j-1}{K-1} \, \bar{\sigma}, \qquad \sigma_K = 0,
\]
so heterogeneity arises from progressively larger perturbation magnitudes, while the last client remains noiseless as a control.

\textbf{Fixed-$k$ label skew.}
Each client $C_j$ is assigned a subset $\mathcal{C}_j \subseteq [K]$ of exactly $k$ distinct labels. The assignment ensures that each label $c \in [K]$ is assigned to at least one client, with balanced coverage across clients.

For each class $c \in [K]$, the index set $\mathcal{I}_c = \{ i \mid y_i = c \}$ is partitioned into $t_c$ shards:
\[
\mathcal{I}_c = \bigcup_{s=1}^{t_c} \mathcal{I}_c^{(s)}, \quad \mathcal{I}_c^{(s)} \cap \mathcal{I}_c^{(s')} = \emptyset \text{ for } s \neq s'.
\]

Each shard $\mathcal{I}_c^{(s)}$ is assigned to a client $C_j$ with $c \in \mathcal{C}_j$. The local dataset of client $C_j$ is then
\[
\mathcal{D}_j = \bigcup_{c \in \mathcal{C}_j} \mathcal{I}_c^{(j,c)}.
\]

\subsection{Datasets and Models}

We evaluate FedTopo on Fashion-MNIST, CIFAR-10 and CIFAR-100 (50k/10k splits; per-dataset mean--std normalization). For Fashion-MNIST we use a shallow CNN; for CIFAR-10/100 we adopt a CIFAR-style ResNet-18. The topology-informative block for TE extraction is selected automatically by TGBS (default: \texttt{layer2} on ResNet-18; \texttt{conv1} on the CNN), consistent with Section~\ref{appendix:tgbc}.

\paragraph{ResNet-18 for CIFAR (CIFAR-style stem).}
The model follows a standard 3$\times$3 stem (stride~1, padding~1) and four residual stages with BasicBlocks; the first block in stages \texttt{layer2--4} uses stride~2 and a 1$\times$1 projection for downsampling. Global average pooling and a linear classifier complete the network. A compact specification is given in Table~\ref{tab:C2-resnet18}.%
\footnote{Each BasicBlock has two 3$\times$3 conv--BN--ReLU layers; when \texttt{stride}$=2$, a 1$\times$1 projection with matching stride is used. BN: \texttt{eps}$=10^{-5}$, \texttt{momentum}$=0.1$.}

\begin{table}[htbp]
\centering
\caption{ResNet-18 (CIFAR variant) architecture summary for $32\times32$ RGB input.}
\label{tab:C2-resnet18}
\begin{tabular}{lcccc}
\toprule
Stage & Operator (kernel/stride) & Channels & Output size & Notes \\
\midrule
\texttt{conv1} & Conv2d $3\times3$/1 & 64  & $32\times32$ & BN, ReLU \\
\texttt{layer1} & $2\times$ BasicBlock $(3\times3$/1)$\times2$ & 64  & $32\times32$ & No downsample \\
\texttt{layer2} & $[$BasicBlock $(3\times3$/2)$\times2] \times 1$; then $(3\times3$/1)$\times2$ & 128 & $16\times16$ & First block uses 1$\times$1 proj. \\
\texttt{layer3} & BasicBlock $(3\times3$/2)$\times2$; then $(3\times3$/1)$\times2$ & 256 & $8\times8$ & First block uses 1$\times$1 proj. \\
\texttt{layer4} & BasicBlock $(3\times3$/2)$\times2$; then $(3\times3$/1)$\times2$ & 512 & $4\times4$ & First block uses 1$\times$1 proj. \\
\texttt{avgpool} & AdaptiveAvgPool2d $(1\times1)$ & -- & $1\times1$ & -- \\
\texttt{fc} & Linear & $K$ & $1\times1$ & $K{=}$ \#classes (10/100) \\
\bottomrule
\end{tabular}
\end{table}

\paragraph{Simple CNN for Fashion-MNIST.}
The CNN follows the provided configuration (two conv layers with a single pooling layer shared symbolically as \texttt{pool}, three fully-connected layers). For a $28\times28$ grayscale input, the tensor sizes are summarized in Table~\ref{tab:C2-simplecnn}.%
\footnote{The listing corresponds to \texttt{SimpleCNNMNIST}: \texttt{conv1}$\to$\texttt{pool}$\to$\texttt{conv2}$\to$\texttt{fc1}$\to$\texttt{fc2}$\to$\texttt{fc3}. The pooling is MaxPool2d, kernel$=2$, stride$=2$.}

\begin{table}[htbp]
\centering
\caption{Simple CNN (Fashion-MNIST) architecture summary for $28\times28$ grayscale input.}
\label{tab:C2-simplecnn}
\begin{tabular}{lcccc}
\toprule
Layer & Operator (kernel/stride) & Channels & Output size & Notes \\
\midrule
\texttt{conv1} & Conv2d $5\times5$/1 & 6   & $24\times24$ & -- \\
\texttt{pool}  & MaxPool2d $2\times2$/2 & 6   & $12\times12$ & -- \\
\texttt{conv2} & Conv2d $5\times5$/1 & 16  & $8\times8$ & -- \\
\texttt{pool}  & MaxPool2d $2\times2$/2 & 16  & $4\times4$ & -- \\
\texttt{flatten} & -- & 16$\times$4$\times$4 & 256 & -- \\
\texttt{fc1} & Linear & 120 & -- & ReLU \\
\texttt{fc2} & Linear & 84  & -- & ReLU \\
\texttt{fc3} & Linear & 10  & -- & Classifier \\
\bottomrule
\end{tabular}
\end{table}

\paragraph{Remark on the provided instantiations.}
The above tables compactly reflect the detailed PyTorch printouts you supplied. In your CIFAR runs, the printed \texttt{fc} sometimes shows \texttt{out\_features=1000} (ImageNet-style head). In all experiments we set the classifier to the dataset's class count ($K{=}10$ or $100$); the topology module (TGBS/TE/TAL) is agnostic to this choice because it operates on the intermediate block (default \texttt{layer2} for ResNet-18; \texttt{conv1} for the CNN).

\subsection{Training Protocol and Hyperparameters}

Unless otherwise stated, clients train locally with mini-batch SGD and periodically synchronize with the server. Table~\ref{tab:C-train-hparams} summarizes the common training protocol and the topology-specific settings used across experiments. Topological features are computed per channel using 2-D cubical sublevel filtrations on activation maps; persistence diagrams are mapped to persistence images (PIs) and averaged to form the TE. The adaptive weight $\alpha$ follows the schedules defined in Appendix~\ref{app:alpha-scheduling}.

\begin{table}[htbp]
    \centering
    \caption{Training protocol and hyperparameters (defaults and commonly used values).}
    \label{tab:C-train-hparams}
    \begin{tabular}{ll}
        \toprule
        \textbf{Protocol (FL)} & \\
        \midrule
        Clients ($n$)                    & 5 \\
        Communication rounds             & 10 (CIFAR); 5 (FMNIST, unless noted) \\
        Local epochs per round ($E$)     & 5 \\
        Batch size                       & 32 \\
        Optimizer                        & SGD (momentum $0.9$) \\
        Learning rate                    & 0.01 \\
        LR scheduler                     & StepLR (step\_size $=30$, $\gamma=0.01$) \\
        Model                            & CNN (FMNIST); ResNet-18 (CIFAR-10/100) \\
        TGBS-selected block              & By screening; default ResNet-18:\ \texttt{layer2}, CNN:\ \texttt{conv1} \\
        \midrule
        \textbf{Topology module (TE/TAL)} & \\
        \midrule
        Filtration on feature maps       & 2-D cubical sublevel (per channel) \\
        Homology degrees                 & $H_0$, $H_1$ \\
        PI vector length ($M$)           & 64 (unless otherwise stated) \\
        Samples per client for TE        & 64 per round (typical) \\
        TE aggregation                   & Channel-wise PI averaging \\
        \midrule
        \textbf{Adaptive $\alpha$ scheduling} & \\
        \midrule
        Schedule family                  & Warm-up / Linear-Topo / Piecewise / Smooth-Topo \\
        Base settings (common)           & $\alpha_{\text{init}}=0.2$, window$=3$, $\gamma_{\alpha}=0.05$ \\
        Linear-/Smooth-Topo ranges       & $\alpha_{\max}\in[0.6,0.8]$, client-specific $\alpha_{\min}^{(i)}$ by imbalance \\
        Normalization range              & $[\ell_{\min},\ell_{\max}]$ tuned per dataset (typical FMNIST $[0.1,0.85]$, CIFAR $[0.5,0.75]$) \\
        \midrule
        \textbf{Visualization} & \\
        \midrule
        UMAP projection                  & 2-D (default; 3-D used for supplementary views when noted) \\
        \bottomrule
    \end{tabular}
\end{table}

\subsection{Feature Block Selection for FedTopo} \label{appendix:feature-block-selection}

FedTopo applies topological regularization to a selected intermediate block. The selection is done using TGBS:

\begin{itemize}
    \item For CNN: features are extracted from \texttt{conv1}.
    \item For ResNet-18: features are extracted from \texttt{layer2}.
\end{itemize}

This choice maximizes topological separability, as validated in the ablation study.

\section{Additional Experimental Results} \label{appendix:additional-results}

\setcounter{figure}{0}
\setcounter{table}{0}
\setcounter{equation}{0}
\renewcommand{\thefigure}{D.\arabic{figure}}
\renewcommand{\thetable}{D.\arabic{table}}
\renewcommand{\theequation}{D.\arabic{equation}}

\begin{figure}[htbp]
    \centering
    \includegraphics[width=\linewidth]{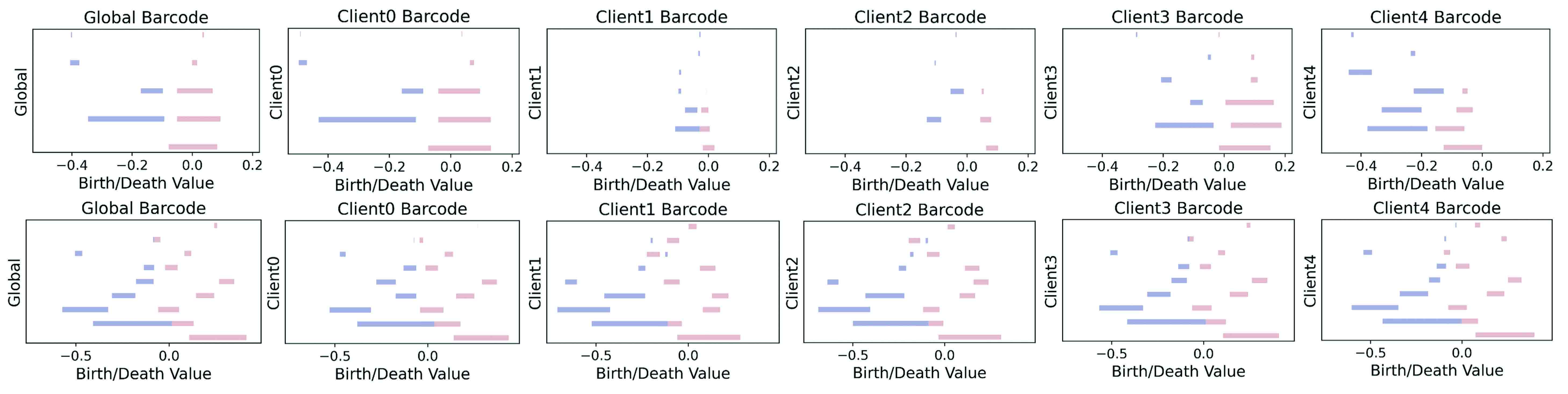}
    \caption{Persistence barcodes of the global model and each client. Each row shows the birth/death values for H$_0$ (connected components, blue) and H$_1$ (holes, red) features \emph{before} (Round~0) and \emph{after} training (Round~9). At Round~0, clients exhibit heterogeneous topologies: H$_0$ contains many short-lived bars (fragmented components) and H$_1$ lifetimes are scattered. By Round~9, client barcodes converge toward the global model’s profile, with a marked reduction of short H$_0$ bars (more compact components) and tighter, similarly located H$_1$ bands (more consistent hole structure). This progression indicates that TAL encourages clients to adopt a common topological signature, corroborating the UMAP trends in Figure~\ref{fig:umap_proj}.}
    \label{fig:barcode_panel}
\end{figure}

This appendix complements the UMAP analysis in the main text by providing \emph{topology-space} evidence of alignment. Specifically, we visualize persistence barcodes computed from the TGBS-selected block for the global model and all clients at the beginning and the end of training (Figure~\ref{fig:barcode_panel}). Whereas the UMAP projections in Figure~\ref{fig:umap_proj} reflect low-dimensional geometric overlap, the barcodes shown here summarize multi-scale homological structure (H$_0$/H$_1$) directly on the feature maps and are thus insensitive to downstream projection artifacts. Consistency between these two views strengthens the claim that TAL drives cross-client alignment in representation \emph{and} topology.

\noindent \textbf{Observations.} (i) \emph{Round~0 heterogeneity:} Across clients, H$_0$/H$_1$ patterns differ substantially in both number and lifetime, evidencing non-I.I.D.\ feature topology. (ii) \emph{Round~9 alignment:} Client barcodes become highly similar to one another and to the global model, with H$_0$ simplification (fewer, longer-lived components) and stabilized H$_1$ lifetimes. (iii) \emph{Implication:} Together with the UMAP convergence (Figure~\ref{fig:umap_proj}), these barcode trends support that TAL reduces representation drift by aligning the \emph{topological embedding} across clients.

\section{Recent Advances in Federated Learning} \label{appendix:related-work}

Federated learning (FL) enables collaborative training across decentralized clients while preserving data privacy~\shortcite{iqua2025federated}. However, performance degrades markedly under non-I.I.D.\ data due to representational drift and unstable convergence~\shortcite{mdpi2025federated}. A large body of work seeks to mitigate these issues, often by improving alignment between local and global models or by enhancing optimization stability~\shortcite{iqua2025federated}. We group representative approaches into five categories: \textbf{optimization-based methods}, \textbf{data augmentation and synthesis}, \textbf{knowledge transfer}, \textbf{robust learning frameworks}, and \textbf{representation alignment \& personalization}. This appendix consolidates and extends the discussion from the main text’s related work, providing a structured summary under non-I.I.D.\ regimes.

\subsubsection{Optimization-Based Methods}
These methods modify the client or server objectives/updates to stabilize training and cope with statistical heterogeneity.
\begin{itemize}
    \item \textbf{FedAvg}~\shortcite{mcmahan2017communication}: The foundational parameter-averaging algorithm. Effective in i.i.d.\ settings, but susceptible to client-drift under non-I.I.D.\ data.
    \item \textbf{FedProx}~\shortcite{li2020federated}: Adds a proximal term to local objectives, constraining updates around the global model to improve stability with heterogeneous data.
    \item \textbf{SCAFFOLD}~\shortcite{karimireddy2020scaffold}: Uses \emph{control variates} to correct client drift by reducing variance in local gradients, accelerating convergence under heterogeneity.
    \item \textbf{FedDyn}~\shortcite{jin2023feddyn}: Introduces \emph{dynamic regularization} on clients, adapting the objective to counteract inconsistency across updates and improve convergence.
    \item \textbf{FedDC}~\shortcite{gao2022feddc}: Decouples local drift from the global objective and applies explicit \emph{drift correction}, yielding faster, more stable training across clients.
\end{itemize}

\subsubsection{Data Augmentation and Synthesis Approaches}
These techniques generate or transform data (or features) to reduce distribution gaps and improve robustness.
\begin{itemize}
    \item \textbf{SDA-FL}~\shortcite{Li2023FederatedGAN}: Employs \emph{GAN}-based synthesis to mimic the global distribution, alleviating non-I.I.D.\ effects during training.
    \item \textbf{FRAug}~\shortcite{Chen2023FRAug}: Performs \emph{representation augmentation} so clients produce personalized embeddings that still align across participants.
    \item \textbf{FedFTG}~\shortcite{zhang2022fine}: Uses \emph{data-free knowledge distillation} and pseudo-sample generation to bridge client–server distribution shifts.
    \item \textbf{MuPFL}~\shortcite{Zhang2024MultiLevel}: Combines \emph{adaptive clustering} with \emph{prior-assisted fine-tuning} to handle long-tailed and non-I.I.D.\ data, improving local and global performance.
\end{itemize}

\subsubsection{Knowledge Transfer Techniques}
These methods transfer knowledge via distillation or contrastive objectives to promote cross-client consistency.
\begin{itemize}
    \item \textbf{FedDF}~\shortcite{lin2020ensemble}: Aggregates client knowledge through \emph{distillation}, reducing reliance on raw data while mitigating heterogeneity.
    \item \textbf{FedGen}~\shortcite{zhu2021data}: Trains a \emph{generator} at the server to produce pseudo-samples approximating the global distribution for improved alignment.
    \item \textbf{MOON}~\shortcite{li2021model}: Applies \emph{model-level contrastive learning} to encourage consistent representations between local and global models under distribution shift.
\end{itemize}

\subsubsection{Robust Learning Frameworks}
These frameworks emphasize robustness (e.g., to adversaries, severe skew, or client failures) in non-I.I.D.\ settings.
\begin{itemize}
    \item \textbf{Cominer}~\shortcite{Zou2022Efficient}: Combines \emph{label clustering} and \emph{vertical comparison} to detect malicious clients and improve robustness and accuracy.
    \item \textbf{FGT}~\shortcite{Li2023GradientCalibration}: Calibrates conflicting gradients across clients to enhance convergence and end-task performance under heterogeneity.
    \item \textbf{FedCSPC}~\shortcite{Qi2023CrossSilo}: Uses \emph{cross-silo prototypical calibration} to create shared prototypes that reduce feature-space heterogeneity across data centers.
    \item \textbf{DReS-FL}~\shortcite{Shao2022DReSFL}: Leverages \emph{secret data sharing} and \emph{Lagrange encoding} for secure, robust learning with malicious clients or dropouts.
    \item \textbf{FedRANE}~\shortcite{Liao2023JointLocal}: Couples \emph{local relational augmentation} with a \emph{global Nash equilibrium} objective to reduce local bias and align with the global model.
    \item \textbf{HyperFed}~\shortcite{Liao2023HyperFed}: Learns \emph{hyperbolic prototypes} with consistent aggregation to capture hierarchical relations and improve stability under non-I.I.D.\ data.
    \item \textbf{FedCD}~\shortcite{Long2023FedCD}: Addresses \emph{class imbalance} and \emph{classifier bias} via fine-grained prototypes, global distillation, and adaptive class-level aggregation.
\end{itemize}

\subsubsection{Representation Alignment and Personalization}
A complementary thread directly targets representation drift by aligning features, classifiers, or semantics across clients.
\begin{itemize}
    \item \textbf{FedCCFA}~\shortcite{neurips2024fedccfa}: \emph{Clusters client classifiers} and \emph{aligns intermediate features} to mitigate concept drift, improving cross-client consistency under non-I.I.D.\ data.
    \item \textbf{FedSA}~\shortcite{arxiv2025fedsa}: Introduces \emph{semantic anchors} that regularize representation learning and curb feedback loops induced by statistical and model heterogeneity.
\end{itemize}

\paragraph{Takeaway.} Across these directions, a common theme is to \emph{stabilize optimization} and \emph{align representations} to counter non-I.I.D.\ effects. Our approach, \textsc{FedTopo}, advances this line by aligning clients in a \emph{topological} representation space, complementing pixel/patch or purely geometric objectives with multi-scale, stable topological descriptors.

\end{document}